\documentclass[11pt]{article}

\usepackage{amssymb,amsmath,amsthm,sectsty,url}
\usepackage[letterpaper,hmargin=1.0in,vmargin=1.0in]{geometry}
\usepackage[pdftex,colorlinks,linkcolor=blue,citecolor=blue,filecolor=blue,urlcolor=blue]{hyperref}
\usepackage{cleveref}
\usepackage{color}
\usepackage[boxed]{algorithm}
\usepackage{tikz}
\usepackage{graphicx}
\usepackage{nicefrac}
\usepackage{aliascnt}
\usepackage{subfig}
\usepackage{booktabs} % for professional tables
\usepackage[boxed]{algorithm}
\usepackage{algorithmic}

\newtheorem{theorem}{Theorem}[section]
\crefname{theorem}{Theorem}{Theorems}

\newaliascnt{lemma}{theorem}

\aliascntresetthe{lemma}
\crefname{lemma}{Lemma}{Lemmas}

\newaliascnt{proposition}{theorem}

\aliascntresetthe{proposition}
\crefname{proposition}{Proposition}{Propositions}

\newaliascnt{corollary}{theorem}

\aliascntresetthe{corollary}
\crefname{corollary}{Corollary}{Corollaries}

\newaliascnt{fact}{theorem}

\aliascntresetthe{fact}
\crefname{fact}{Fact}{Facts}

\newaliascnt{definition}{theorem}

\aliascntresetthe{definition}
\crefname{definition}{Definition}{Definitions}

\newaliascnt{remark}{theorem}

\aliascntresetthe{remark}
\crefname{remark}{Remark}{Remarks}

\newaliascnt{conjecture}{theorem}

\aliascntresetthe{conjecture}
\crefname{conjecture}{Conjecture}{Conjectures}

\newaliascnt{claim}{theorem}

\aliascntresetthe{claim}
\crefname{claim}{Claim}{Claims}

\newaliascnt{question}{theorem}

\aliascntresetthe{question}
\crefname{question}{Question}{Questions}

\newaliascnt{exercise}{theorem}

\aliascntresetthe{exercise}
\crefname{exercise}{Exercise}{Exercises}

\newaliascnt{example}{theorem}

\aliascntresetthe{example}
\crefname{example}{Example}{Examples}

\newaliascnt{notation}{theorem}

\aliascntresetthe{notation}
\crefname{notation}{Notation}{Notations}

\newaliascnt{problem}{theorem}

\aliascntresetthe{problem}
\crefname{problem}{Problem}{Problems}

\def\E{\mathbb E}
\newcommand{\R}{\mathbb R}

\newcommand\Rbb{\ensuremath{\mathbb{R}}}

\title{Learning Space Partitions for Nearest Neighbor Search}

\author{
  Yihe Dong\footnote{Author names are ordered alphabetically.}\\
  Microsoft \\
  \and
  Piotr Indyk \\
  MIT \\
  \and
  Ilya Razenshteyn \\
  Microsoft Research \\
  \and
  Tal Wagner \\
  MIT \\
}

\newenvironment{CompactEnumerate}{
\begin{list}{\arabic{enumi}.}{%
\usecounter{enumi}
\setlength{\leftmargin}{14pt}
\setlength{\itemindent}{1pt}
\setlength{\itemsep}{1pt}
}}
{\end{list}}

\newenvironment{CompactItemize}{
\begin{list}{$\bullet$}{%
\setlength{\leftmargin}{14pt}
\setlength{\itemindent}{0pt}
\setlength{\itemsep}{0pt}
}}
{\end{list}}

\begin{document}

\maketitle

\begin{abstract}
Space partitions of $\mathbb{R}^d$ underlie a vast and important
class of fast nearest neighbor search (NNS) algorithms.
Inspired by recent theoretical work on NNS for general metric spaces~\cite{andoni2018data,andoni2018holder}, we develop a new framework for building space partitions reducing the problem
to \emph{balanced graph partitioning} followed by
\emph{supervised classification.}
We instantiate this general approach with the KaHIP graph partitioner~\cite{sandersschulz2013} and neural networks,
respectively,
to obtain a new partitioning procedure called \emph{Neural Locality-Sensitive Hashing (Neural LSH).}
On several standard benchmarks for NNS~\cite{aumuller2017ann}, our experiments show that the partitions obtained by Neural LSH consistently outperform partitions found by quantization-based and tree-based methods as well as classic, data-oblivious LSH.
\end{abstract}

\section{Introduction}

%\irnote{Add a sentence about scaling.}

The {\it Nearest Neighbor Search (NNS)} problem is defined as follows. Given an $n$-point dataset $P$ in a $d$-dimensional Euclidean space $\mathbb{R}^d$, we would like to preprocess $P$ to answer $k$-nearest neighbor queries quickly.
That is, given a query point $q \in \mathbb{R}^d$, we want to find the $k$ data points from $P$
that are closest to $q$. NNS is a cornerstone of the modern data analysis and, at the same time, a fundamental geometric data structure problem that led to many exciting theoretical developments over the past decades. See, e.g., \cite{wang2016learning,andoni2018approximate} for an overview.

The main two approaches to constructing efficient NNS data structures are \emph{indexing} and \emph{sketching}. The goal of indexing is to construct a data structure that, given a query point, produces a small subset of $P$ (called  \emph{candidate set}) that includes the desired neighbors. Such a data structure can be stored on a single machine,  or (if the data set is very large) distributed among  multiple machines. In contrast, the goal of sketching is to compute compressed representations  of points to enable computing approximate distances quickly (e.g., compact binary hash codes with the Hamming distance used as an estimator, see the surveys~\cite{wang2014hashing,wang2016learning}). 
Indexing and sketching can be (and often are) combined to maximize the overall performance \cite{wu2017multiscale,johnson2017billion}.

Both indexing and sketching have been the topic of a vast amount of theoretical and empirical literature.
In this work, we consider the \emph{indexing} problem. In particular, we focus on indexing  based on \emph{space partitions}.
The overarching idea is to build a partition of the ambient space $\mathbb{R}^d$ and split the dataset $P$ accordingly.
Given a query point  $q$, we identify the bin containing $q$ and
form the resulting list of candidates from the data points residing in the same bin (or, to boost the accuracy, nearby bins as well).
%To boost the search accuracy, it is often necessary to add all data points from  \emph{nearby} bins to the candidate list
%(this is often referred as the \emph{multi-probe} technique).
%The literature about space partitions for NNS is so vast that it would be completely futile to try to give a thorough overview. Instead,
%we list the main classes of methods that have been showing great performance: 
Some of the popular space partitioning methods include locality-sensitive hashing (LSH)~\cite{lv2007multi,andoni2015practical,dasgupta2017neural}; quantization-based approaches,
where partitions are obtained via $k$-means clustering of the dataset~\cite{jegou2011product,babenko2012inverted}; and tree-based methods such as  random-projection trees or PCA trees~\cite{sproull1991refinements,bawa2005lsh,dasgupta2013randomized, keivani2018improved}.

Compared to other indexing methods, space partitions have multiple benefits. First, they are naturally applicable in {\em distributed} settings, as different bins can be stored on different machines~\cite{bahmani2012efficient, nidetecting,  li2017losha, bhaskara2018distributed}. Moverover, the computational efficiency of search can be further improved by using any nearest neighbor search algorithm locally on each machine. Second, partition-based indexing is particularly suitable for GPUs due to the simple and predictable memory access pattern~\cite{johnson2017billion}. Finally, partitions can be combined with cryptographic techniques to yield efficient {\em secure} similarity search algorithms~\cite{chen2019sanns}. Thus, in this paper we focus on designing space partitions that optimize the trade-off between their key metrics: the number of reported candidates, the fraction of the true nearest neighbors among the candidates, the number of bins, and the computational efficiency of the point location.

Recently, there has been a large body of work that studies how modern machine learning
techniques (such as neural networks) can help tackle various classic algorithmic problems (a partial list includes~\cite{mousavi2015deep,baldassarre2016learning,bora2017compressed,khalil2017learning,metzler2017learned,kraska2017case, balcan2018learning,lykouris2018competitive,mitz2018model,purohit2018improving}).
Similar methods---under the name ``learn to hash''---have been used to improve the \emph{sketching} approach to NNS~\cite{wang2016learning}. 
%However, when it comes to \emph{indexing}, only rather simple unsupervised techniques such as PCA or $k$-means have been (successfully) used.
%\irnote{This might have triggered the reviewer who mentioned learned kd-trees. I think we need to reformulate this sentence.}\twnote{Suggestion:}
However, when it comes to~\emph{indexing}, while some unsupervised techniques such as PCA or $k$-means have been successfully applied, the full power of modern tools like neural networks has not yet been harnessed. This state of affairs naturally
leads to the following general question:
    \textbf{Can we employ modern (supervised) machine learning techniques
    to find good space partitions for nearest neighbor search?}

%\irnote{The following paragraph needs to be fixed!}
%Recent work has shown that supervised machine learning methods, such as neural networks, can be used to obtain improved solutions for various classic algorithmic problems, including nearest neighbor search~\cite{wang2016learning,kraska2017case, balcan2018learning,lykouris2018competitive,mitz2018model}. However, even though these methods have been applied to the  \emph{sketching} approach  (more on this below), we are not aware of any successful attempts to do the same for the \emph{indexing} approach.
%This state of affairs naturally leads to the following question: 
%
%\begin{center}
%    \textbf{Can we employ (supervised) machine learning
%    techniques
%    to find good space partitions for nearest neighbor search?}
%\end{center}

\subsection{Our contribution}

In this paper we address the aforementioned challenge and present a new framework for finding high-quality space partitions of $\mathbb{R}^d$.
Our approach consists of three major steps:
\begin{CompactEnumerate}
\item Build the $k$-NN graph $G$ of the dataset by connecting each data point to $k$ nearest neighbors;
\item Find a balanced partition $\mathcal{P}$ of the graph $G$ into $m$ parts of nearly-equal size such that the number of edges between different parts is as small as possible;
\item Obtain a partition of $\mathbb{R}^d$ by training a classifier on the data points with labels being the parts of the partition $\mathcal{P}$ found in the second step.
\end{CompactEnumerate}

\begin{figure*}%
    \centering
    \subfloat[Dataset]{{\includegraphics[width=0.23\textwidth]{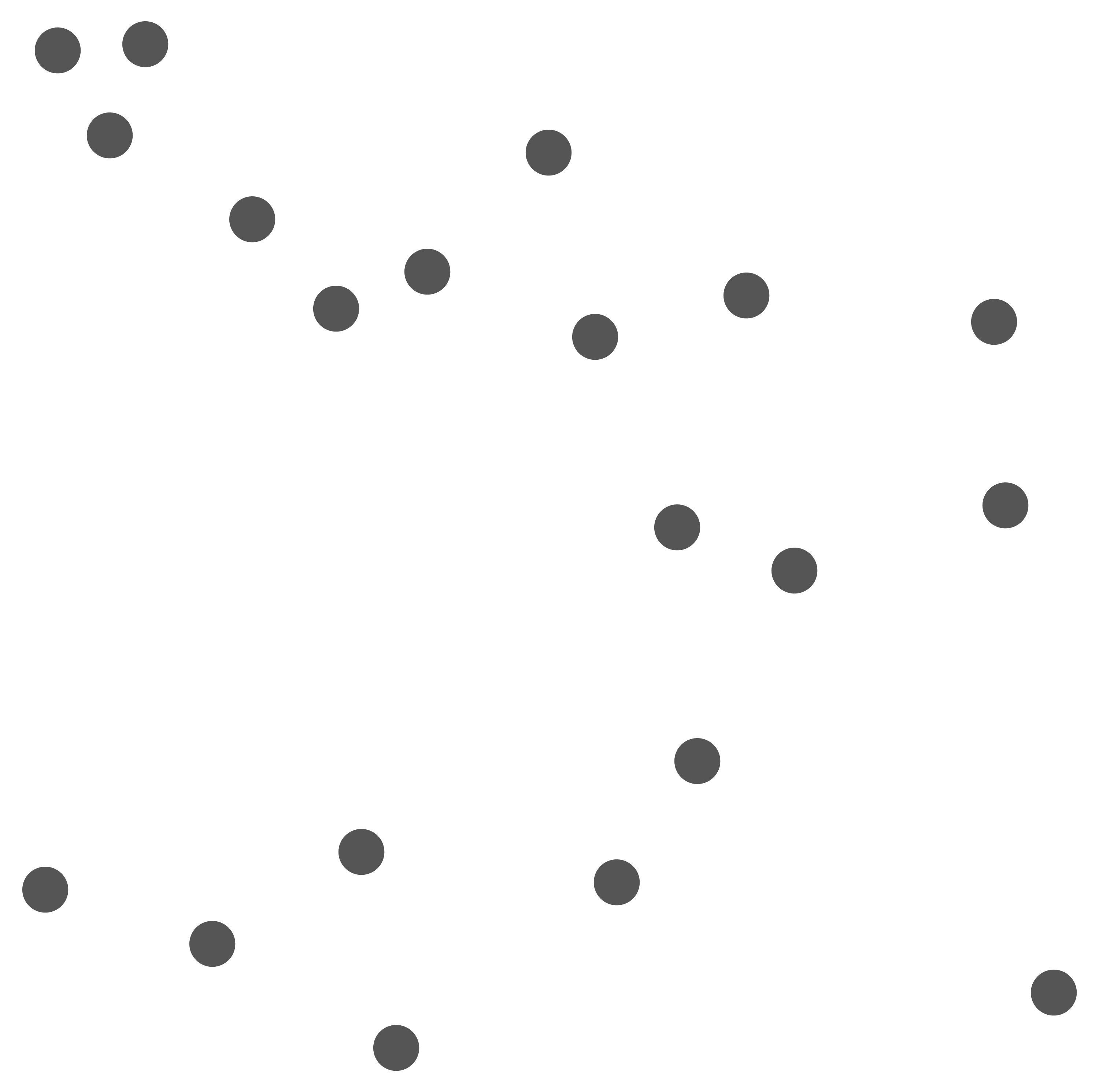}}}%
    \hfill
    \subfloat[$k$-NN graph together with a balanced partition]{{\includegraphics[width=0.23\textwidth]{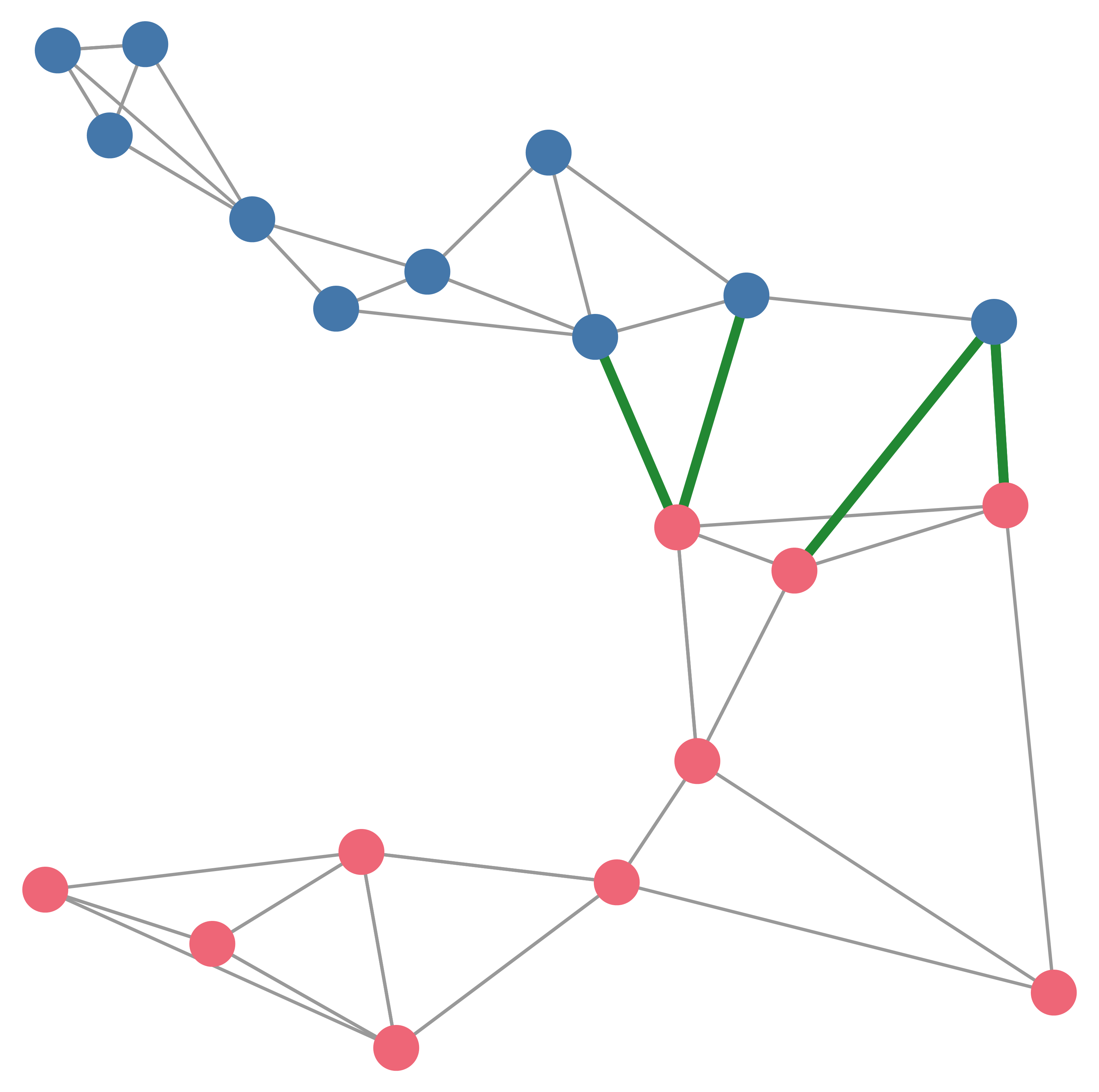}}}%
    \hfill
    \subfloat[Learned partition]{{\includegraphics[width=0.23\textwidth]{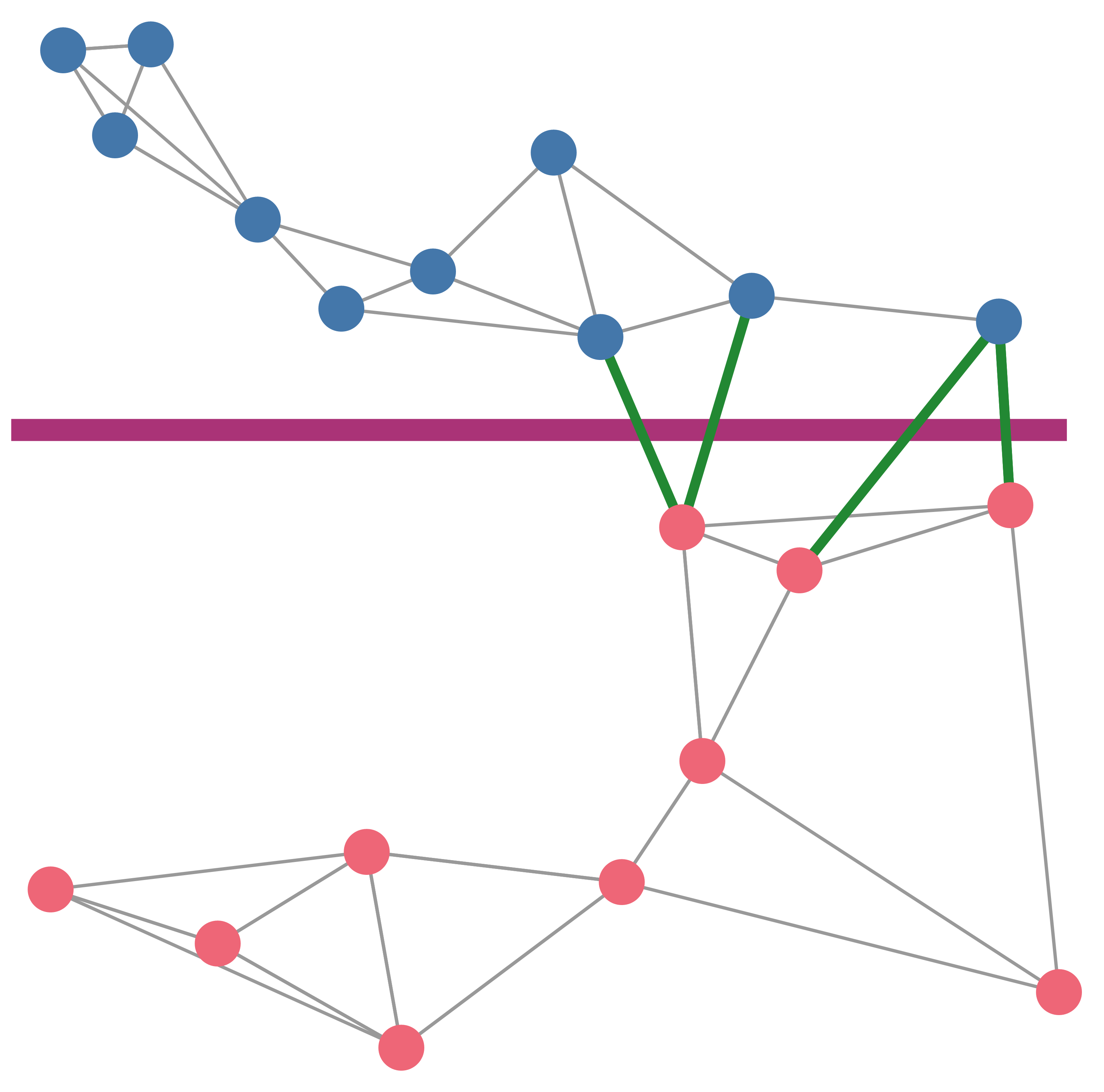}}}%
    \caption{Stages of our framework}
        \label{fig:pipeline}
\end{figure*}

See Figure~\ref{fig:pipeline} for illustration.
The new algorithm \emph{directly optimizes} the performance of the partition-based nearest neighbor data structure.
Indeed, if a query is chosen as a uniformly random \emph{data point,} then the average $k$-NN accuracy is exactly equal to the fraction of edges of the $k$-NN
graph $G$ whose endpoints are separated by the partition $\mathcal{P}$. This generalizes to out-of-sample queries provided that
the query and dataset distributions are close, and the test accuracy of the trained classifier is high.

At the same time, our approach is directly related to and inspired by recent theoretical work~\cite{andoni2018data,andoni2018holder} on NNS for general metric spaces.
In particular, using the framework of~\cite{andoni2018data,andoni2018holder}, we prove that, under mild conditions on the dataset $P$, the $k$-NN graph of $P$
can be partitioned with a hyperplane into two parts of comparable size such that only few edges get split by the hyperplane. This gives a partial theoretical justification of our method.

The new framework is very flexible and uses partitioning and learning in a black-box way. This 
    allows us to plug various models (linear models, neural networks,
    etc.) and explore the trade-off between the quality
    and the algorithmic efficiency of the resulting partitions.
We emphasize the importance of~\emph{balanced} partitions for the indexing problem, where all bins contain roughly the same number of data points.
This property is crucial in the distributed setting, since
we naturally would like to assign a similar number
of points to each machine. 
Furthermore, balanced partitions allow tighter control
of the number of candidates simply by varying the number of retrieved parts.
Note that a priori, it is unclear how to partition $\mathbb{R}^d$ so as to induce balanced bins of a given dataset. Here the combinatorial portion of our approach is particularly useful, as balanced graph partitioning is a well-studied problem, and our supervised extension to $\mathbb{R}^d$ naturally preserves the balance by virtue of attaining high training accuracy.

We speculate that the new method might be potentially useful for solving the NNS problem for \emph{non-Euclidean} metrics, such as the edit distance~\cite{zhang2017embedjoin} or optimal transport distance~\cite{kusner2015word}. Indeed, for any metric space,
one can compute the $k$-NN graph and then partition it. The only step
that needs to be adjusted to the specific metric at hand is the learning step.

Let us finally put forward the challenge of scaling our method up to billion-sized or
even
larger datasets. For such scale, one needs
to build an \emph{approximate} $k$-NN
graph as well as using graph partitioning algorithms
that are faster than KaHIP. We leave
this exciting direction to future work. For the current experiments (datasets of size $10^6$ points), preprocessing takes several hours.
Another important challenge is to obtain NNS algorithms
based on the above partitioning with \emph{provable}
guarantees in terms of approximation and running time. However, we expect it to be difficult,
in particular, since all the current state-of-the-art NNS algorithms lack such guarantees~(e.g., $k$-means-based~\cite{jegou2011product} or graph methods~\cite{malkov2018efficient}, see also \cite{aumuller2017ann} for a recent SOTA survey).

\paragraph{Evaluation} We instantiate our framework with the KaHIP algorithm~\cite{sandersschulz2013} for the graph partitioning step, and either linear models or small-size neural networks for the learning step.
We evaluate it on several standard benchmarks for NNS~\cite{aumuller2017ann} and conclude that in terms of quality of the resulting partitions, it consistently outperforms quantization-based and tree-based partitioning
procedures, while maintaining comparable algorithmic efficiency. In the high accuracy regime,
our framework yields partitions that lead to processing up to $2.3\times$ fewer candidates than the strongest baseline.

As a baseline method we use $k$-means clustering~\cite{jegou2011product}. It produces a partition of the dataset into $k$ bins, in a way that naturally extends to all of $\mathbb{R}^d$, by assigning a query point $q$ to its nearest centroid. (More generally, for multi-probe querying, we can rank the bins by the distance of their centroids to $q$). This simple scheme yields very high-quality results for indexing. Besides $k$-means, we evaluate LSH~\cite{andoni2015practical}, ITQ~\cite{gong2013iterative}, PCA tree~\cite{sproull1991refinements}, RP tree~\cite{dasgupta2013randomized}, and Neural Catalyzer~\cite{sablayrolles2018spreading}.

\subsection{Related work}\label{sec:related}
On the empirical side, currently the fastest indexing techniques for the NNS problem are \emph{graph-based}~\cite{malkov2018efficient}. The high-level idea is to construct a graph on the dataset
(it can be the $k$-NN graph, but other constructions are also possible), and then for each query perform a walk, which eventually converges to the nearest neighbor.
Although very fast, graph-based approaches have suboptimal ``locality of reference'', which makes them less suitable for several modern architectures. For instance, this is the case when the algorithm is run on a GPU~\cite{johnson2017billion}, or when the data is stored in external memory~\cite{sun2014srs} or in a distributed manner~\cite{bahmani2012efficient,nidetecting}.
Moreover, graph-based indexing requires many rounds of adaptive access to the dataset, whereas partition-based indexing accesses the dataset in one shot.
This is crucial, for example, for nearest neighbor search over encrypted data~\cite{chen2019sanns}.
These benefits justify further study of partition-based methods.

Machine learning techniques are particularly useful for the \emph{sketching} approach, leading to a vast body of research 
under the label ``learning to hash''~\cite{wang2014hashing,wang2016learning}.
In particular, several recent works employed neural networks to obtain high-quality sketches~\cite{erin2015deep,sablayrolles2018spreading}.
The fundamental difference from our work is that sketching is designed to speed up~\emph{linear scans} over the dataset, by reducing the \emph{cost} of distance evaluation, while indexing is designed for~\emph{sublinear time} searches, by reducing the \emph{number} of distance evaluations.
We note that while sketches are not designed for indexing, they can be used for that purpose, since a $b$-bit hashing scheme induces a partition of $\R^d$ into $2^b$ parts.
Nonetheless, our experiments show that partitions induced by these methods (such as Iterative Quantization~\cite{gong2013iterative}) are not well-suited for indexing, and underperform compared to quantization-based indexing, as well as to our methods.

We highlight in particular the recent work of~\cite{sablayrolles2018spreading}, which uses neural networks to learn a mapping $f \colon \mathbb{R}^d \to \mathbb{R}^{d'}$
that improves the geometry of the dataset and the queries to facilitate subsequent sketching.
%It is natural to apply the same family of maps for \emph{partitions}; however,
It is natural to ask whether the same family of maps can be applied to enhance the quality of \emph{partitions} for indexing. However,
as our experiments show, in the high accuracy regime the maps learned using the algorithm of~\cite{sablayrolles2018spreading}
consistently degrade the quality of partitions.

%using partitions obtained this way for indexing is very different from the intended
%use of sketches, and we confirm experimentally that high-quality sketching techniques such as ITQ~\cite{gong2013iterative} produce partitions
%that underperform compared to quantization-based partitions as well as the framework we introduce in this paper.

%A different application of neural networks related to NNS is to optimize the performance of the nearest neighbor~\emph{classifier}.
%Given a labeled dataset in a classification setting, the idea is to learn a representation of the dataset -- either as sketches~\cite{klein2017defense,jain2017subic} or as a high-dimensional embedding~\cite{schmidt2018learning} -- that would render the nearest neighbor classifier (i.e., labeling each query point with the label of its nearest data point) as accurate as possible. Apart from not producing an indexing method, these works are also different from ours by being inherently supervised, relying on a fully labeled dataset, whereas our approach is unsupervised.

Finally, we mention that here is some prior work on learning space partitions: \cite{cayton2008learning,ram2013space,li2011learning}.
However, all these algorithms learn \emph{hyperplane} partitions into two parts (then applying them recursively).
Our method, on the other hand, is much more flexible, since neural networks allow us to learn a much richer class of partitions.

\section{Our method}

%\paragraph{Training}
Given a dataset $P\subseteq\mathbb{R}^d$ of $n$ points, and a number of bins $m>0$, our goal is to find a partition $\mathcal{R}$ of $\mathbb{R}^d$ into $m$ bins with the following properties:
\begin{CompactEnumerate}
    \item\emph{Balanced:} The number of data points in each bin is not much larger than $n / m$.
    \item\emph{Locality sensitive:} For a typical query point $q \in \mathbb{R}^d$, most of its nearest neighbors belong to the same bin of $\mathcal{R}$. We assume that queries and data points come from similar distributions.
    \item\emph{Simple:} The partition should admit a compact description
    and, moreover, the point location process should be computationally efficient. For example, we might look for a space partition induced by hyperplanes.
\end{CompactEnumerate}

Formally, we want the partition $\mathcal{R}$ that minimizes the loss $\E_q\left[\sum_{p\in N_k(q)}\mathbf1_{\mathcal{R}(p)\neq\mathcal{R}(q)}\right]$ s.t.~$\forall_{p\in P}\;|\mathcal{R}(p)|\leq (1+\eta)(n/m)$,
%\[ \E_q\left[\sum_{p\in N_k(q)}\mathbf1_{\mathcal{R}(p)\neq\mathcal{R}(q)}\right] \;\;\;\; s.t. \;\;\;\; \forall_{p\in P}\;|\mathcal{R}(p)|\leq O(m/n) , \]
where $q$ is sampled from the query distribution, $N_k(q)\subset P$ is the set of its $k$ nearest neighbors in $P$, $\eta>0$ is a balance parameter, and $\mathcal{R}(p)$ denotes the part of $\mathcal{R}$ that contains $p$.

First, suppose that the query is chosen as a \emph{uniformly random data point}, $q \sim P$. Let $G$ be the $k$-NN graph of $P$, whose vertices are the data points, and each vertex
is connected to its $k$ nearest neighbors. Then the above problem boils down to partitioning vertices of the graph $G$ into $m$ bins such that each bin contains roughly $n / m$ vertices, and the number of edges crossing between different bins is as small as possible (see Figure~\ref{fig:pipeline}(b)). This \emph{balanced graph partitioning}
problem is extremely well-studied, and there are available combinatorial partitioning solvers that produce very high-quality solutions.
In our implementation, we use the open-source solver KaHIP~\cite{sandersschulz2013}, which is based on a sophisticated local search.

More generally, we need to handle out-of-sample queries, i.e., which are not contained in $P$.
Let $\widetilde{\mathcal{R}}$ denote the partition of $G$ (equivalently, of the dataset $P$) found by the graph partitioner.
To convert $\widetilde{\mathcal{R}}$ into a solution to our problem, we need to extend it to a partition $\mathcal{R}$ of the whole space $\mathbb{R}^d$ that would work well for query points.
In order to accomplish this, we train a model that, given a query point $q \in \mathbb{R}^d$,
predicts which of the $m$ bins of $\widetilde{\mathcal{R}}$ the point $q$ belongs to (see Figure~\ref{fig:pipeline}(c)). We use the dataset $P$ as a training set,
and the partition $\widetilde{\mathcal{R}}$ as the labels -- i.e., each data point is labeled with the ID of the bin of $\widetilde{\mathcal{R}}$ containing it. 
The method is summarized in Algorithm~\ref{alg:main}.
The geometric intuition for this learning step is that -- even though the partition $\widetilde{\mathcal{R}}$ is obtained by combinatorial means, and in principle might consist of ill-behaved subsets of $\mathbb{R}^d$ -- in most practical scenarios, we actually expect it to be close to being induced by a simple partition of the ambient space.
For example, if the dataset
is fairly well-distributed on the unit sphere, %$S^{d-1} \subset \mathbb{R}^d$
and the number of bins is $m = 2$,
a balanced cut of $G$ should be close to a hyperplane.

The choice of model to train depends on the desired properties of the output partition $\mathcal{R}$.
For instance, if we are interested in a hyperplane partition,
we can train a linear model using SVM or regression.  In this paper, we instantiate the learning step with both \emph{linear models}
and \emph{small-sized neural networks.}
Here, there is natural tension between the size of the model we train and the accuracy of the resulting classifier, and hence the quality of the partition we produce.
A larger model yields better NNS accuracy, at the expense of computational efficiency.
We discuss this in Section 3.

\paragraph{Multi-probe querying}
Given a query point $q$, the trained model can be used to assign it to a bin of a partition $\mathcal{R}$, and search for nearest neighbors within the data points in that part.
In order to achieve high search accuracy,
we actually train the model to predict \emph{several} bins for a given query point,
which are likely to contain nearest neighbors.
For neural networks, this can
be done naturally by taking several largest outputs of the last layer.
By searching through more bins (in the order of preference predicted by the model) we can achieve better accuracy, allowing for a trade-off between computational resources and accuracy.

\begin{figure}
    \centering
    \includegraphics[width=0.47\textwidth]{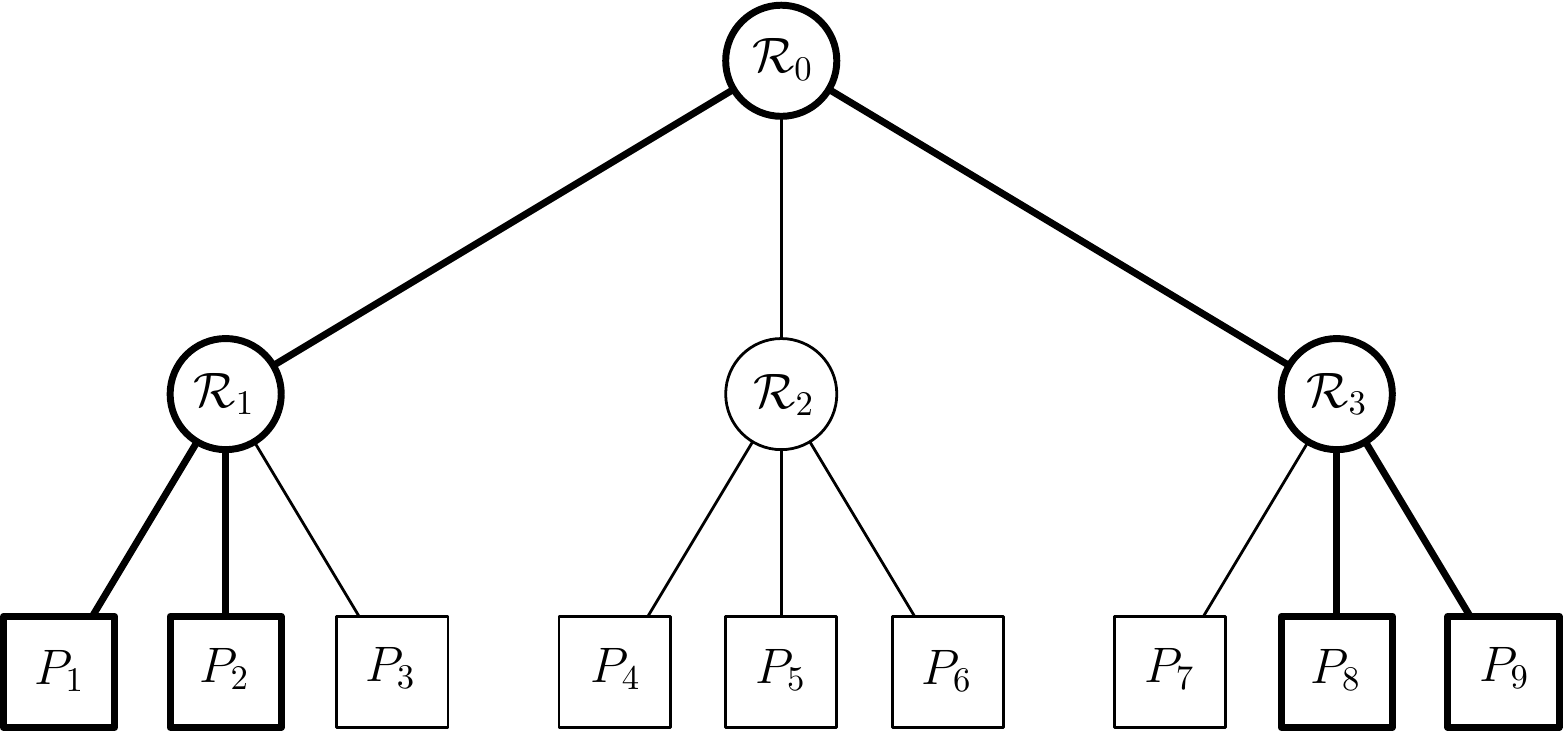}
    \caption{Hierarchical partition into $9$ bins with $m_1 = m_2 = 3$. $\mathcal{R}_i$'s are partitions,
    $P_j$'s are the bins of the dataset. Multi-probe query procedure, which descends into $2$ bins, may visit the bins marked in bold.}
        \label{fig:hierarchical}
\end{figure}

\paragraph{Hierarchical partitions} When the required number of bins $m$ is large, in order to improve the efficiency of the resulting partition,
it pays off to produce it in a hierarchical manner. Namely, we first find a partition of $\mathbb{R}^d$ into $m_1$ bins, then recursively partition each
of the bins into $m_2$ bins, and so on, repeating the partitioning for $L$ levels.
The total number of bins in the overall partition is $m = m_1 \cdot m_2 \cdot \ldots m_L$.
See Figure~\ref{fig:hierarchical} for illustration.
The advantage of such a hierarchical partition is that it is much simpler to navigate
than a one-shot partition with $m$ bins.

\newcommand{\INDENT}{\hspace{1em}}
\begin{algorithm}[!t]
\caption{Nearest neighbor search with a learned space partition}
\label{alg:main}
%\smallskip{\hrule height.2pt}\smallskip
\smallskip
\textbf{Preprocessing}\\ 
Input: Dataset $P\subset\R^d$, integer parameter $k>0$, number of bins $m>0$
\smallskip{\hrule height.2pt}\smallskip
\smallskip
\begin{algorithmic}[1]
   \STATE Build a $k$-NN graph $G$ of $P$.
   \STATE Run a balanced graph partitioning algorithm on $G$ into $m$ parts. Number the parts arbitrarily as $1,\ldots,m$. Let $\pi(p)\in\{1,\ldots,m\}$ denote the part containing $p$, for every $p\in P$.
   \STATE Train a machine learning model $M$ with training set $P$ and labels $\{\pi(p)\}_{p\in P}$. For every $x\in\R^d$, let $M(x)\in\{1,\ldots,m\}$ denote the prediction of $M$ on $x$.
\smallskip
\end{algorithmic}
\smallskip
$M(\cdot)$ defines our $m$-way partition of $\R^d$. Note that it is possible that $\pi(p)\neq M(p)$ for some $p\in P$, if $M$ attains imperfect training accuracy.
\smallskip{\hrule height.2pt}\smallskip
\smallskip
\textbf{Query}\\ 
Input: query point $q\in\R^d$, number of bins to search $b$
\smallskip{\hrule height.2pt}\smallskip
\smallskip
\begin{algorithmic}[1]
   \STATE Run inference on $M$ to compute $M(q)$.
   \STATE Search for a near neighbor of $q$ in the bin $M(q)$, i.e., among the candidates $\{p\in P:M(p)=M(q)\}$.
   \STATE If $M$ furthermore predicts a~\emph{distribution} over bins, search for a near neighbor in the $b$ top-ranked bins according to the ranking induced by the distribution (i.e., from the most likely bin to less likely ones).
\smallskip
\end{algorithmic}
\end{algorithm}

%\subsection{Neural LSH}
\paragraph{Neural LSH with soft labels}
In the primary instantiation of our framework, we set the supervised learning component to a a neural network with a small number of layers and constrained hidden dimensions (the exact parameters are specified in the next section).
%In one instantiation of the supervised learning component, we use neural networks with a small number of layers and constrained hidden dimensions. The exact parameters are specified in the next section.
%The exact parameters depend on the size of the training set, and are specified in the next section.
%
%The depths of the networks and the hidden dimension are intentionally constrained, to reduce overfitting, memory usage, and training time.
%
%\irnote{Soft labels!}
%
%\paragraph{Soft labels}
In order to support effective multi-probe querying, we need to infer not just the bin that contains the query point, but rather a \emph{distribution} over bins that are likely to contain this point and its neighbors.
A $T$-probe candidate list is then formed from all data points in the $T$ most likely bins.
In order to accomplish this, we use~\emph{soft labels} for data points generated as follows. For $S \geq 1$ and a data point $p$, the soft label
$\mathcal{P} = (p_1, p_2, \ldots, p_m)$ is a distribution over the bin containing a point chosen uniformly at random among $S$ nearest neighbors of $p$ (including $p$ itself).
Now, for a predicted distribution $\mathcal{Q}=(q_1, q_2,\ldots,q_m)$, we seek to minimize the KL divergence between $\mathcal{P}$ and $\mathcal{Q}$:
$\sum_{i=1}^m p_i \log \frac{p_i}{q_i}$.
Intuitively, soft labels help guide the neural network with information about multiple bin ranking. $S$ is a hyperparameter that needs to be tuned; we study its setting in the appendix (cf.~Figure~\ref{fig:hyper_plots_s}).
%Section~\ref{sec:additional_experiments}.
%We study its setting in Section~\ref{sec:additional_experiments}.

%The purpose of the soft labels is to guide the neural network with information about the ranking of bins for searching nearest neighbors. Optimizing w.r.t.~$\mathcal{P}$ allows the model to predict multiple bins more accurately, which is necessary for achieving high accuracy via multi-probe querying.
%
%$S$ is a hyperparameter that needs to be tuned. In practice, the accuracy in the objective function increases in the regime where $S$ is larger than $k$, as more neighbors give the network a more ``complete'' distribution over bins.

\section{Sparse hyperplane-induced cuts in $k$-NN graphs}

We state and prove a theorem that shows, under certain mild assumptions, that the $k$-NN graph of a dataset $P \subseteq \Rbb^d$ can be partitioned
by a hyperplane such that the induced cut is sparse (i.e., has few crossing edges while the sizes of two parts are similar).
The theorem is based on the framework of~\cite{andoni2018data,andoni2018holder}
and uses spectral techniques.

We start with some notation.
Let $N_k(p)$ be the set of $k$ nearest neighbors of $p$ in $P$.
The degree of $p$ in the $k$-NN graph is $\deg(p) = |N_k(p)\cup\{p' \in P \mid p \in N_k(p')\}|$.
Let $\mathcal{D}$ be the distribution over the dataset $P$, where a point $p \in P$ is sampled with probability proportional to its degree $\deg(p)$.
Let $\mathcal{D}_{\mathrm{close}}$ be the distribution over pairs $(p, p') \in P \times P$,
where $p \in P$ is uniformly random, and $p'$ is a uniformly random element of $N_k(p)$. 
%(the set of $k$ nearest neighbors of $p$).
Denote $\alpha = \mathrm{E}_{(p, p') \in \mathcal{D}_{\mathrm{close}}}[\|p - p'\|_2^2]$
and $\beta = \mathrm{E}_{x_1 \sim \mathcal{D}, x_2 \sim \mathcal{D}}[\|p_1 - p_2\|_2^2]$.
We will proceed assuming that $\alpha$ (typical distance between a data point and its
nearest neighbors) is noticeably smaller than $\beta$ (typical distance between two
independent data points).

The following theorem implies, informally speaking, that if
$\alpha \ll \beta$, then there exists a hyperplane
which splits the dataset into two parts of not too different size
while separating only few pairs of $(p, p')$, where $p'$ is
one of the $k$ nearest neighbors of $p$.
For the proof of the theorem, see Appendix~\ref{spectral_proof}.

\begin{theorem}
\label{spectral_thm}
There exists a hyperplane $H = \{x \in \Rbb^d \mid \langle a, x\rangle = b\}$ such that the following holds.
Let $P = P_1 \cup P_2$ be the partition of $P$ induced by $H$: $P_1 = \{p \in P \mid \langle a, p \rangle \leq b\}$,
$P_2 = \{p \in P \mid \langle a, p \rangle > b\}$. Then, one has:
\begin{equation}
\label{eqeqeq3}
\frac{\mathrm{Pr}_{(p, p') \sim \mathcal{D}_{\mathrm{close}}}[\mbox{$p$ and $p'$ are separated by $H$}]}{\min\{\mathrm{Pr}_{p \sim \mathcal{D}}[p \in P_1],\mathrm{Pr}_{p \sim \mathcal{D}}[p \in P_2]\}} \leq \sqrt{\frac{2 \alpha}{\beta}}.
\end{equation}
\end{theorem}

\section{Experiments}\label{sec:experiments}

\paragraph{Datasets}
For the experimental evaluation, we use three standard ANN benchmarks~\cite{aumuller2017ann}: SIFT (image descriptors, 1M 128-dimensional points), GloVe (word embeddings~\cite{pennington2014glove}, approximately 1.2M 100-dimensional points, normalized), and MNIST (images of digits, 60K 784-dimensional points).
All three datasets come with $10\,000$ query points, which are used for evaluation. We include the results for SIFT and GloVe in the main text, and MNIST in Appendix~\ref{appendix_mnist}.

\paragraph{Evaluation metrics}
We mainly investigate the trade-off between the number of candidates generated for a query point, and the $k$-NN accuracy, defined as the fraction of its $k$ nearest neighbors that are among those candidates.
The number of candidates determines the processing time of an individual query.
Over the entire query set, we report both the \emph{average}
as well as the \emph{$0.95$-th quantile} of the number of candidates.
The former measures the \emph{throughput}\footnote{Number of queries per second.} of the data structure, while the latter measures its \emph{latency.}\footnote{Maximum time per query, modulo a small fraction of outliers.}
We focus on parameter regimes that yield $k$-NN accuracy of at least $0.75$, in the setting $k=10$.
Additional results with broader regimes of accuracy and of $k$ are included in the appendix.
%We mostly focus on parameter regimes that lead to $k$-NN accuracy of at least $0.8$.
%The experiments in this section use $k = 10$; see appendix for other values of $k$.

\paragraph{Our methods}
We evaluate two variants of our method, with two different choices of the supervised learning component:% in our framework.

\begin{CompactItemize}
\item\textbf{Neural LSH:}
In this variant we use small neural networks.
%Their exact architecture is detailed in the next section.
%We compare Neural LSH to partitions obtained by $k$-means clustering. As mentioned in Section 1, this method produces high quality partitions, and the other methods we evaluate (LSH and
%and Iterative Quantization) do not match it.
We compare this method with $k$-means clustering, Iterative Quantization (ITQ)~\cite{gong2013iterative}, Cross-polytope LSH~\cite{andoni2015practical}, and Neural Catalyzer~\cite{sablayrolles2018spreading} composed over $k$-means clustering.
We evaluate partitions into $16$ bins and $256$ bins. 
We test both one-level (non-hierarchical) and two-level (hierarchical) partitions.
Queries are multi-probe.

\item\textbf{Regression LSH:} This variant uses logistic regression as the supervised learning component
and, as a result, produces very simple partitions induced by \emph{hyperplanes.}
We compare this method with PCA trees~\cite{sproull1991refinements,kumar2008good,abdullah2014spectral}, random projection trees~\cite{dasgupta2013randomized}, and recursive bisections using $2$-means clustering.
We build trees of hierarchical bisections of depth up to $10$ (thus total number of leaves up to $1024$).
The query procedure descends a single root-to-leaf path and returns the candidates in that leaf.
\end{CompactItemize}

\subsection{Implementation details}

Neural LSH uses a fixed neural network architecture for the top-level partition,
and a fixed architecture for all second-level partitions.
Both architectures consist of several blocks, where each block is a fully-connected layer + batch normalization~\cite{ioffe2015batch} + ReLU activations.
The final block is followed by a fully-connected layer and a softmax layer.
The resulting network predicts a distribution over the bins of the partition.
The only difference between the top-level network the second-level network architecture is their number of blocks ($b$) and the size of their hidden layers ($s$).
In the top-level network we use $b=3$ and $s=512$.
In the second-level networks we use $b=2$ and $s=390$.
%
%The only differences between the two architectures are the number of blocks and the sizes of hidden layers: for the top-level partition, we have three blocks and the hidden layers are of size $512$, while for the second-level partitions, one has two blocks, and the size of the hidden layers is $390$.
%
To reduce overfitting, we use dropout with probability $0.1$ during training.
The networks are trained using the Adam optimizer \cite{adam2015} for under $20$ epochs on both levels. We reduce the learning rate multiplicatively at regular intervals. The weights are initialized with Glorot initialization~\cite{glorot2010}.
To tune soft labels, we try different values of $S$ between $1$ and $120$.

We evaluate two settings for the number of bins in each level, $m=16$ and $m=256$ (leading to a total number of bins of the total number of bins in the two-level experiments are $16^2=256$ and $256^2 = 65\,536$, respectively).
In the two-level setting with $m=256$ the bottom level of Neural LSH uses $k$-means instead of a neural network, to avoid overfitting when the number of points per bin is tiny. The other configurations (two-levels with $m=16$ and one-level with either $m=16$ or $m=256$) we use Neural LSH at all levels.

We slightly modify the KaHIP partitioner to make it more efficient on the $k$-NN graphs. Namely, we introduce a hard threshold
of $2000$ on the number of iterations for the local search part of the algorithm, which speeds up the partitioning dramatically,
while barely affecting the quality of the resulting partitions.

\subsection{Comparison with multi-bin methods}\label{sec:kmeanscomparison}
\begin{figure*}%
    \centering
    \subfloat[GloVe, one level, $16$ bins]{{\includegraphics[width=0.50\textwidth]{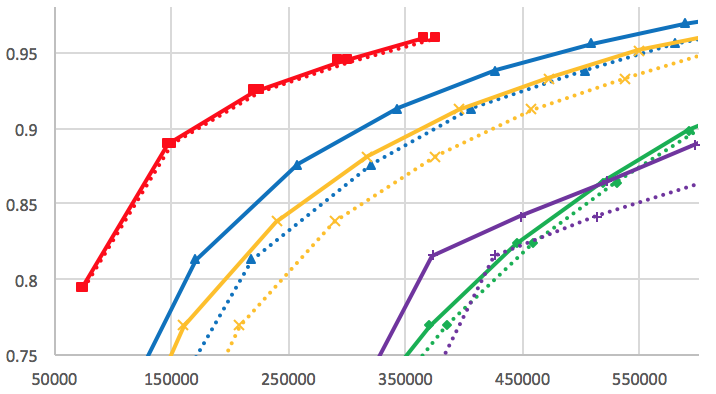}}}%
    \subfloat[SIFT, one level, $16$ bins]{{\includegraphics[width=0.50\textwidth]{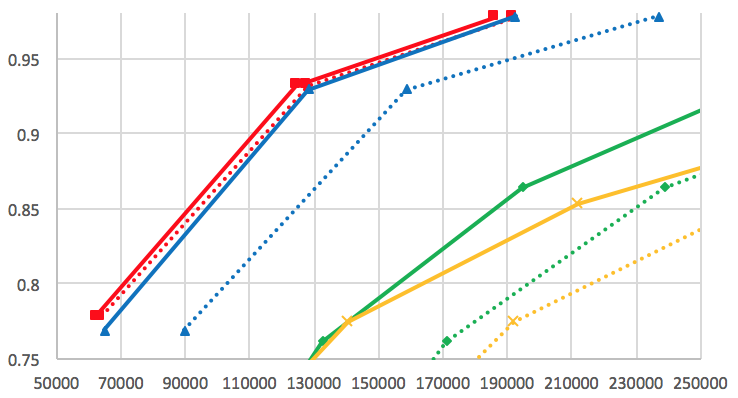}}}%

    \subfloat[GloVe, one level, $256$ bins]{{\includegraphics[width=0.50\textwidth]{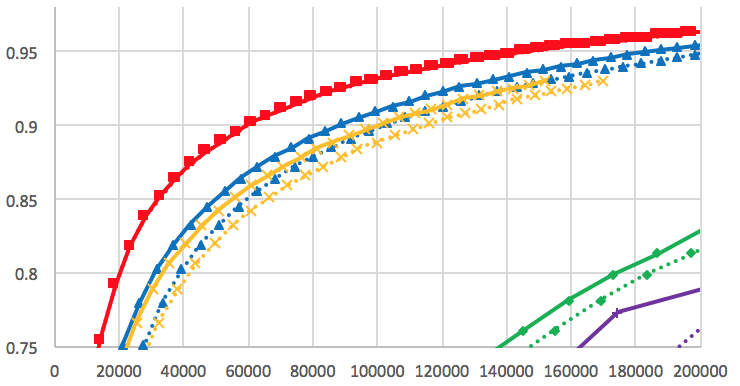}}}%
    \subfloat[SIFT, one level, $256$ bins]{{\includegraphics[width=0.50\textwidth]{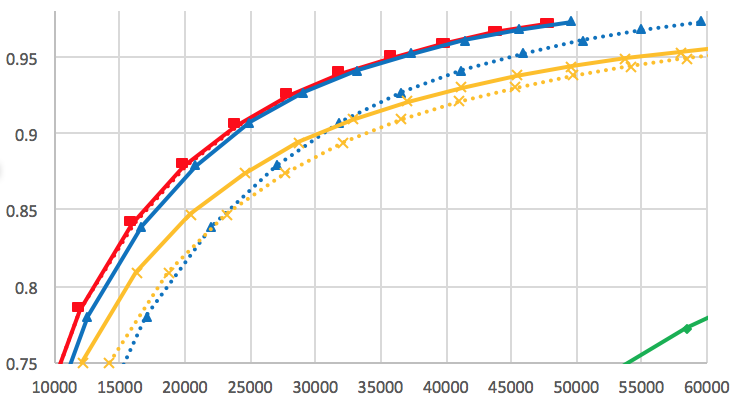}}}%
    
    \subfloat[GloVe, two levels, $16$ bins]{{\includegraphics[width=0.50\textwidth]{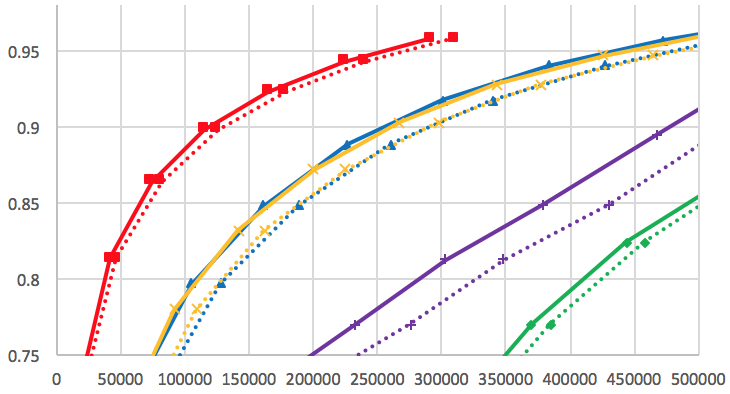}}}%
    \subfloat[SIFT, two levels, $16$ bins]{{\includegraphics[width=0.50\textwidth]{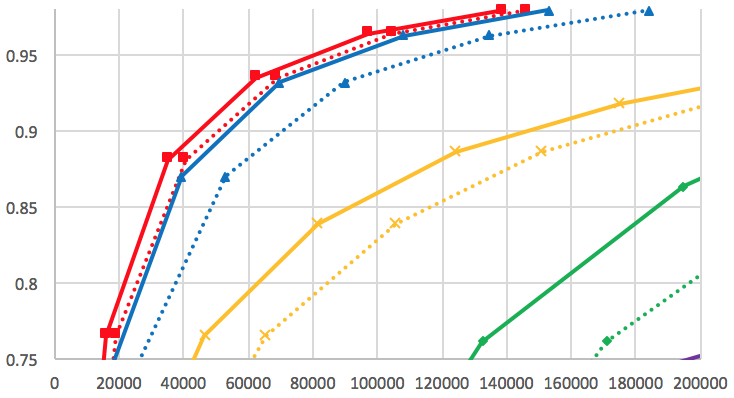}}}%

    \subfloat[GloVe, two levels, $256$ bins, $k$-means at 2nd level]{{\includegraphics[width=0.50\textwidth]{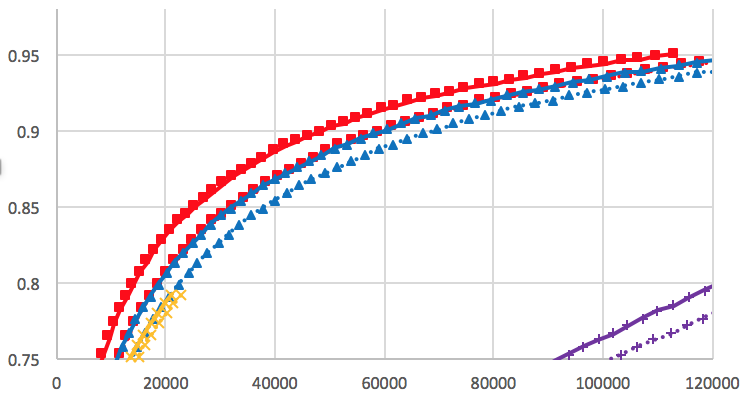}}}%
    \subfloat[SIFT, two levels, $256$ bins, $k$-means at 2nd level]{{\includegraphics[width=0.50\textwidth]{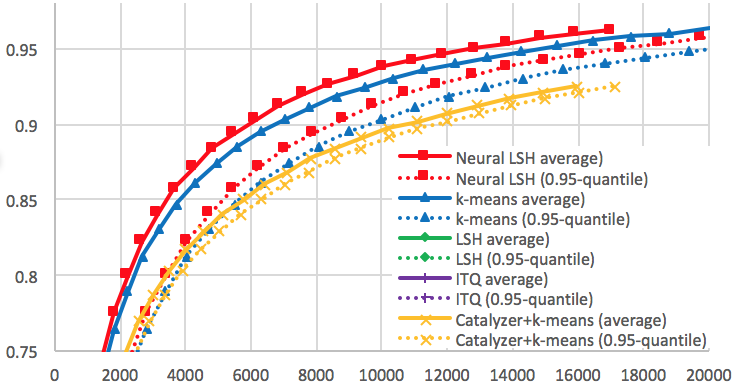}}}%
    \caption{Comparison of Neural LSH with baselines; x-axis is the number of candidates, y-axis is the $10$-NN accuracy}%
    \label{fig:kmeans_plots}
\end{figure*}

Figure~\ref{fig:kmeans_plots} shows the empirical comparison of Neural LSH with $k$-means clustering, ITQ, Cross-polytope LSH, and Neural Catalyzer composed over $k$-means clustering.
It turns out that $k$-means is the strongest among these baselines.\footnote{It is important to note that ITQ is not designed to produce space partitions; as explained in Section 1, it does so as a side-effect. Simiarly, Neural Catalyzer is not designed to enhance partitions. The comparison is intended to show that they do not outperform indexing techniques despite being outside their intended application.} 
% We evaluate these methods outside their intended application.}
The points depicted in Figure~\ref{fig:kmeans_plots} are those that attain accuracy $\geq0.75$. In the appendix (Figure~\ref{fig:all_plots}) we include the full accuracy range for all methods.

%However, if one wishes to use partitions to split points across machines to build a distributed NNS data structure, then a
%single-level settings seems to be more suitable.

In all settings considered, 
Neural LSH yields consistently better partitions than $k$-means.\footnote{We note that two-level partitioning with $m=256$ is the best performing configuration of $k$-means, for both SIFT and GloVe, in terms of the minimum number of candidates that attains $0.9$ accuracy. Thus we evaluate this baseline at its optimal performance.} Depending on the setting, $k$-means requires significantly more candidates to achieve the same accuracy:
\begin{CompactItemize}
    \item Up to $117\%$ more for the average number of candidates for GloVe;
    \item Up to $130\%$ more for the $0.95$-quantiles of candidates for GloVe;
    \item Up to $18\%$ more for the average number of candidates for SIFT;
    \item Up to $34\%$ more for the $0.95$-quantiles of candidates for SIFT;
\end{CompactItemize}
Figure~\ref{table_kmeans} lists the largest multiplicative advantage in the number of candidates of Neural LSH compared to $k$-means, for accuracy values of at least $0.85$.
Specifically, for every configuration of $k$-means, we compute the ratio between the number of candidates in that configuration and the number of candidates of Neural LSH in its optimal configuration, among those that attained at least the same accuracy as that $k$-means configuration.

\begin{figure*}
\centering
\begin{tabular}{|r|r|cc|cc|}
\hline
 && \textbf{GloVe} &  & \textbf{SIFT} & \\
 && Averages & $0.95$-quantiles  & Averages & $0.95$-quantiles \\
 \hline
\textbf{One level} & $16$ bins & 1.745 & 2.125 & 1.031 & 1.240 \\
& $256$ bins & 1.491 & 1.752 & 1.047 & 1.348 \\
\hline
\textbf{Two levels} & $16$ bins & 2.176 & 2.308 & 1.113 & 1.306 \\
& $256$ bins & 1.241 & 1.154 & 1.182 & 1.192 \\
\hline
\end{tabular}
\caption{Largest ratio between the number of candidates for Neural LSH and $k$-means over the settings where both attain the same target $10$-NN accuracy, over accuracies of at least $0.85$.
%for which Neural LSH obtains no worse accuracy than $k$-means.
See details in Section~\ref{sec:kmeanscomparison}.}
\label{table_kmeans}
\end{figure*}

We also note that in all settings except two-level partitioning with $m=256$,\footnote{As mentioned earlier, in this setting Neural LSH uses $k$-means at the second level, due to the large overall number of bins compared to the size of the datasets. This explains why the gap between the average and the $0.95$-quantile number of candidates of Neural LSH is larger for this setting.}
Neural LSH produces partitions for which the 
$0.95$-quantiles for the number of candidates are very close to the average number of candidates, which indicates very little variance between query times over different query points.
In contrast, the respective gap in the partitions produced by $k$-means is much larger, since unlike Neural LSH, it does not directly favor balanced partitions.
This implies that Neural LSH might be particularly suitable for latency-critical
NNS applications.

%We also note that for all the settings except ``two levels, $256$ bins on each'' (where we use $k$-means for the second level) Neural LSH finds partitions, for which the $0.95$-quantiles for the number of candidates are very close to the average number of candidates, however for the $k$-means the respective gap is significant. This disparity leads to the gap between $k$-means and Neural LSH significantly increasing when one passes from averages to the $0.95$-quantiles. Hence, Neural LSH might be particularly suitable for latency-critical NNS applications.

\paragraph{Model sizes.}
The largest model size learned by Neural LSH is equivalent to storing about $\approx 5700$ points for SIFT, or $\approx 7100$ points for GloVe.%\footnote{The difference accounts for the different network architecture used for them, as well as their different dimensionality.}
This is considerably larger than $k$-means with $k\leq256$, which stores at most $256$ points.
Nonetheless, we believe the larger model size is acceptable for Neural LSH, for the following reasons.
%\begin{itemize}
    %\item
    First, in most of the NNS applications, especially for the distributed setting, the bottleneck in the high accuracy regime is the memory accesses needed to retrieve candidates and the further processing (such as distance computations, exact or approximate). The model size is not a hindrance as long as does not exceed certain reasonable limits (e.g., it should fit into a CPU cache).
    Neural LSH significantly reduces the memory access cost, while increasing the model size by an acceptable amount.
    %\item
    Second, we have observed that the quality of the Neural LSH partitions is not too sensitive to decreasing the sizes
    the hidden layers. The model sizes we report are, for the sake of concreteness, the largest ones that still lead to improved performance. Larger models do not increase the accuracy, and sometimes decrease it due to overfitting.
%\end{itemize}

\subsection{Comparison with tree-based methods}

Next we compare binary decision trees, where in each tree node a \emph{hyperplane} is used to determine which of the two subtrees to descend into. We generate hyperplanes with the following methods: Regression LSH, the Learned KD-tree of~\cite{cayton2008learning}, the Boosted Search Forest of~\cite{li2011learning}, cutting the dataset into two equal halves along the top PCA direction~\cite{sproull1991refinements,kumar2008good},
$2$-means clustering, and random projections of the centered dataset~\cite{dasgupta2013randomized,keivani2018improved}.
We build trees of depth up to $10$, which correspond to hierarchical partitions with the up to $2^{10} = 1024$ bins. Results for GloVe and SIFT are summarized in Figure~\ref{fig:trees_results} (see appendix).
For random projections, we run each configuration $30$ times and average the results.

For GloVe, Regression LSH significantly outperforms $2$-means,
while for SIFT, Regression LSH essentially matches $2$-means in terms of the \emph{average} number of candidates,
but shows a noticeable advantage in terms of the $0.95$-percentiles.
In both instances, Regression LSH significantly outperforms PCA tree, and all of the above methods
dramatically improve upon random projections.

Note, however, that random projections have an additional benefit: in order to boost search
accuracy, one can simply repeat the sampling process several times and generate an ensemble of
decision trees instead of a single tree. This allows making each individual tree relatively deep,
which decreases the overall number of candidates, trading space for query time. Other considered approaches (Regression LSH, $2$-means, PCA tree) are inherently deterministic, and boosting their accuracy requires more care:
for instance, one can use partitioning into blocks as in~\cite{jegou2011product}, or alternative approaches like~\cite{keivani2018improved}. Since we focus
on individual partitions and not ensembles, we leave this issue out of the scope.

\begin{figure*}%
    \centering
    \includegraphics[width=\textwidth]{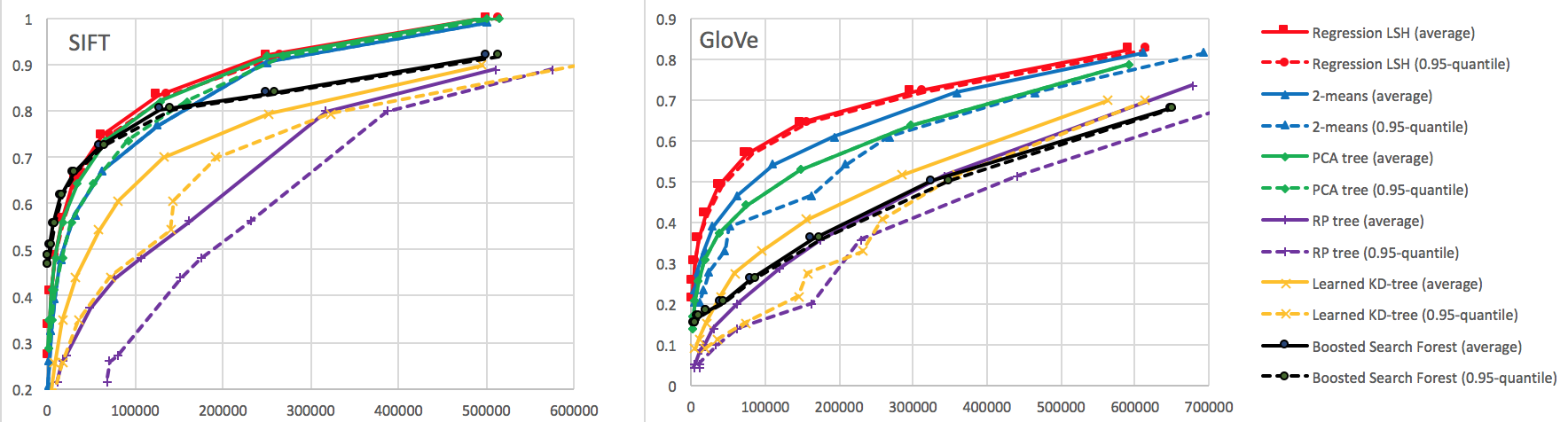}

    \caption{Comparison of decision trees built from hyperplanes: x-axis -- number of candidates, y-axis -- $10$-NN accuracy}%
    \label{fig:trees_results}
\end{figure*}

%\subsection{Additional experiments}\label{sec:additional_experiments}
%\paragraph{Additional experiments.}
%In the appendix we inlclude several additional experiments: \emph{(i)} We evaluate the $50$-NN accuracy of Neural LSH when the partitioning step is run on either the $10$-NN or the $50$-NN graph.\footnote{Neural LSH can solve $k$-NNS by partitioning the $k'$-NN graph, for any $k,k'$; they do not have to be equal.}
%Both settings outperform $k$-means, and the gap
%between using the $10$-NN and $50$-NN graphs is negligible,
%which indicates the robustness of Neural LSH.
%\emph{(ii)} We study the effect of tuning the size of soft labels $S$.
%We show that setting $S$ to be at least $15$ is immensely beneficial
%compared to $S = 1$, but beyond that we start observing diminishing returns.

\subsection{Additional experiments}
In this section we include several additional experiments.

\begin{figure*}%
    \centering
    \subfloat[\label{fig:hyper_plots_k}GloVe, one level, $16$ bins, $50$-NN accuracy using $10$-NN and $50$-NN graphs]{{\includegraphics[width=0.50\textwidth]{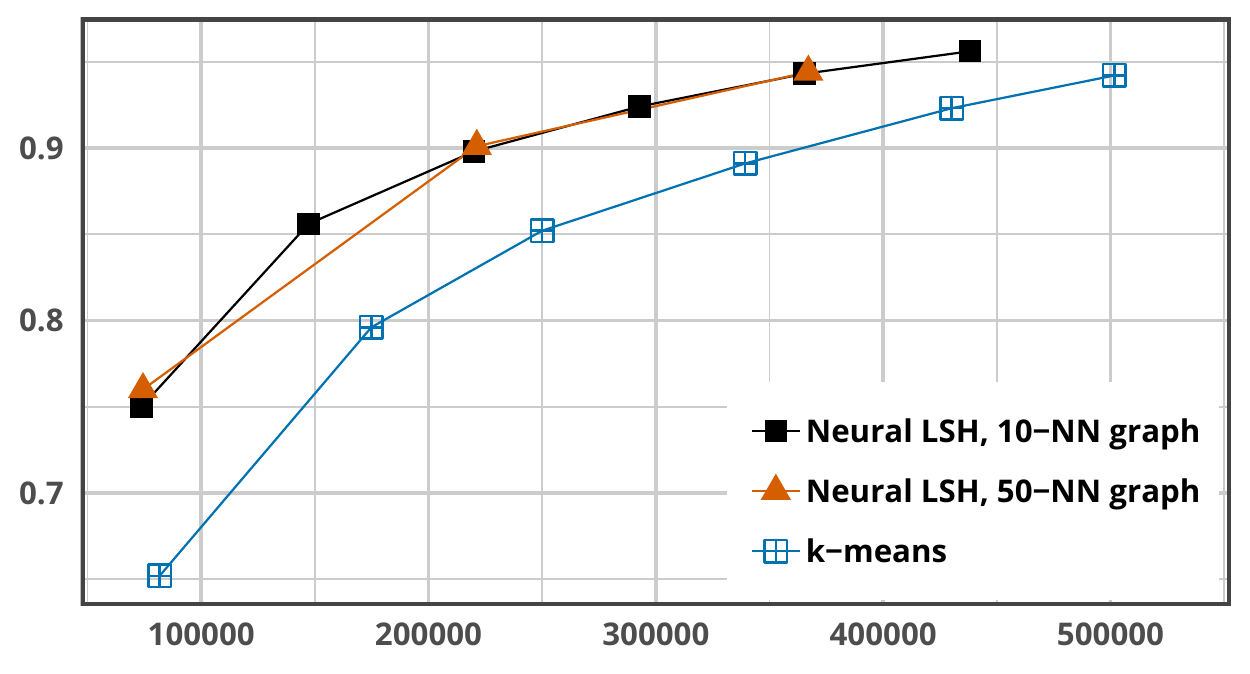}}}%
    \subfloat[\label{fig:hyper_plots_s}GloVe, one level, $256$ bins, varying $S$]{{\includegraphics[width=0.50\textwidth]{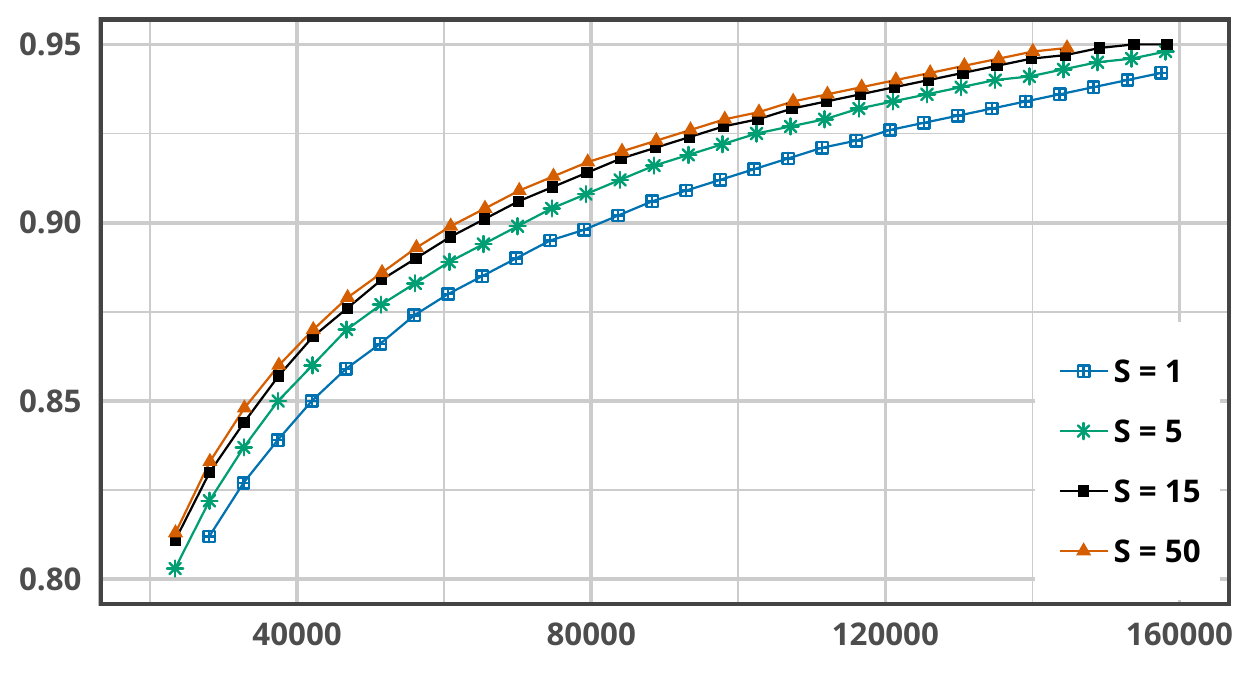}}}%
    \caption{Effect of various hyperparameters}
    \label{fig:hyper_plots}
\end{figure*}

First, we study the effect of setting $k$.
We evaluate the $50$-NN accuracy of Neural LSH when the partitioning step is run on either the $10$-NN or the $50$-NN graph.\footnote{Neural LSH can solve $k$-NNS by partitioning the $k'$-NN graph, for any $k,k'$; they do not have to be equal.} We compare both algorithms to $k$-means with $k = 50$. 
Figure~\ref{fig:hyper_plots_k} compares these three algorithms on
GloVe for $16$ bins reporting
average numbers of candidates. From this plot,
we can see that for $k = 50$, Neural LSH
convincingly outperforms $k$-means,
and whether we use $10$-NN or $50$-NN graph matters very little.

Second, we study the effect of varying $S$ (the soft labels parameter) for Neural LSH on GloVe for $256$ bins.
See Figure~\ref{fig:hyper_plots_s} where we report the average number of candidates. As we can see from the plot,
the setting $S = 15$ yields much better results
compared to the vanilla case of $S = 1$. However,
increasing $S$ beyond $15$ brings diminishing returns  on the overall
accuracy.

\section{Conclusions and future directions}

%In this paper, 
We presented a new technique for finding partitions of $\mathbb{R}^d$ which support high-performance indexing for sublinear-time NNS.
It proceeds in two major steps:
%\begin{CompactItemize}
%    \item We start with combinatorial balanced partitioning of the $k$-NN graph of the dataset;
%    \item We extend the resulting partition to the whole ambient space $\mathbb{R}^d$ by using
%    supervised classification (such as logistic regression, neural networks, etc.).
%\end{CompactItemize}
%
(1) We perform a combinatorial balanced partitioning of the $k$-NN graph of the dataset;
(2) We extend the resulting partition to the whole ambient space $\mathbb{R}^d$ by using
    supervised classification (such as logistic regression, neural networks, etc.).
Our experiments show that the new approach consistently outperforms quantization-based and tree-based
partitions.
There is a number of exciting open problems we would like to highlight:

\begin{CompactItemize}
    \item Can we use our approach for NNS over \emph{non-Euclidean} geometries,
    such as the edit distance~\cite{zhang2017embedjoin} or the optimal transport
    distance~\cite{kusner2015word}?
    The graph partitioning step directly carries through,
    but the learning step may need to be adjusted.
    \item Can we jointly optimize a graph partition \emph{and} 
    a classifier at the same time?
    By making the two components aware of each other,
    we expect the quality of the resulting partition of $\mathbb{R}^d$ to improve.
    A related approach has been successfully applied in~\cite{li2011learning} for hyperplane tree partitions.
    \item Can our approach be extended to learning \emph{several} high-quality partitions that complement each other?
    Such an ensemble might be useful to trade query time for
    memory usage~\cite{andoni2017optimal}.
    \item Can we use machine learning techniques to improve \emph{graph-based} indexing techniques~\cite{malkov2018efficient} for NNS? 
    (This is in contrast to partition-based indexing, as done in this work).
    \item Our framework is an example of combinatorial tools aiding
    ``continuous'' learning techniques.
    A more open-ended question is whether other problems can benefit from such symbiosis.
\end{CompactItemize}

\paragraph{Acknowledgments.}
Supported by NSF TRIPODS awards No.~1740751 and No.~1535851, Simons Investigator Award, and MIT-IBM Watson AI Lab collaboration grant.

\bibliography{bibfile}

\newcommand{\etalchar}[1]{$^{#1}$}
\providecommand{\bysame}{\leavevmode\hbox to3em{\hrulefill}\thinspace}
\providecommand{\MR}{\relax\ifhmode\unskip\space\fi MR }
\providecommand{\MRhref}[2]{%
  \href{http://www.ams.org/mathscinet-getitem?mr=#1}{#2}
}
\providecommand{\href}[2]{#2}
\begin{thebibliography}{ANN{\etalchar{+}}18b}

\bibitem[AAKK14]{abdullah2014spectral}
Amirali Abdullah, Alexandr Andoni, Ravindran Kannan, and Robert Krauthgamer,
  \emph{Spectral approaches to nearest neighbor search}, arXiv preprint
  arXiv:1408.0751 (2014).

\bibitem[ABF17]{aumuller2017ann}
Martin Aum{\"u}ller, Erik Bernhardsson, and Alexander Faithfull,
  \emph{Ann-benchmarks: A benchmarking tool for approximate nearest neighbor
  algorithms}, International Conference on Similarity Search and Applications,
  Springer, 2017, pp.~34--49.

\bibitem[AIL{\etalchar{+}}15]{andoni2015practical}
Alexandr Andoni, Piotr Indyk, Thijs Laarhoven, Ilya Razenshteyn, and Ludwig
  Schmidt, \emph{Practical and optimal lsh for angular distance}, Advances in
  Neural Information Processing Systems, 2015, pp.~1225--1233.

\bibitem[AIR18]{andoni2018approximate}
Alexandr Andoni, Piotr Indyk, and Ilya Razenshteyn, \emph{Approximate nearest
  neighbor search in high dimensions}, arXiv preprint arXiv:1806.09823 (2018).

\bibitem[ALRW17]{andoni2017optimal}
Alexandr Andoni, Thijs Laarhoven, Ilya Razenshteyn, and Erik Waingarten,
  \emph{Optimal hashing-based time-space trade-offs for approximate near
  neighbors}, Proceedings of the Twenty-Eighth Annual ACM-SIAM Symposium on
  Discrete Algorithms, Society for Industrial and Applied Mathematics, 2017,
  pp.~47--66.

\bibitem[ANN{\etalchar{+}}18a]{andoni2018data}
Alexandr Andoni, Assaf Naor, Aleksandar Nikolov, Ilya Razenshteyn, and Erik
  Waingarten, \emph{Data-dependent hashing via nonlinear spectral gaps},
  Proceedings of the 50th Annual ACM SIGACT Symposium on Theory of Computing
  (2018), 787--800.

\bibitem[ANN{\etalchar{+}}18b]{andoni2018holder}
\bysame, \emph{H{\"o}lder homeomorphisms and approximate nearest neighbors},
  2018 IEEE 59th Annual Symposium on Foundations of Computer Science (FOCS),
  IEEE, 2018, pp.~159--169.

\bibitem[BCG05]{bawa2005lsh}
Mayank Bawa, Tyson Condie, and Prasanna Ganesan, \emph{Lsh forest: self-tuning
  indexes for similarity search}, Proceedings of the 14th international
  conference on World Wide Web, ACM, 2005, pp.~651--660.

\bibitem[BDSV18]{balcan2018learning}
Maria-Florina Balcan, Travis Dick, Tuomas Sandholm, and Ellen Vitercik,
  \emph{Learning to branch}, International Conference on Machine Learning,
  2018.

\bibitem[BGS12]{bahmani2012efficient}
Bahman Bahmani, Ashish Goel, and Rajendra Shinde, \emph{Efficient distributed
  locality sensitive hashing}, Proceedings of the 21st ACM international
  conference on Information and knowledge management, ACM, 2012,
  pp.~2174--2178.

\bibitem[BJPD17]{bora2017compressed}
Ashish Bora, Ajil Jalal, Eric Price, and Alexandros~G Dimakis, \emph{Compressed
  sensing using generative models}, International Conference on Machine
  Learning, 2017, pp.~537--546.

\bibitem[BL12]{babenko2012inverted}
Artem Babenko and Victor Lempitsky, \emph{The inverted multi-index}, Computer
  Vision and Pattern Recognition (CVPR), 2012 IEEE Conference on, IEEE, 2012,
  pp.~3069--3076.

\bibitem[BLS{\etalchar{+}}16]{baldassarre2016learning}
Luca Baldassarre, Yen-Huan Li, Jonathan Scarlett, Baran G{\"o}zc{\"u}, Ilija
  Bogunovic, and Volkan Cevher, \emph{Learning-based compressive subsampling},
  IEEE Journal of Selected Topics in Signal Processing \textbf{10} (2016),
  no.~4, 809--822.

\bibitem[BW18]{bhaskara2018distributed}
Aditya Bhaskara and Maheshakya Wijewardena, \emph{Distributed clustering via
  lsh based data partitioning}, International Conference on Machine Learning,
  2018, pp.~569--578.

\bibitem[CCD{\etalchar{+}}19]{chen2019sanns}
Hao Chen, Ilaria Chillotti, Yihe Dong, Oxana Poburinnaya, Ilya Razenshteyn, and
  M~Sadegh Riazi, \emph{Sanns: Scaling up secure approximate k-nearest
  neighbors search}, arXiv preprint arXiv:1904.02033 (2019).

\bibitem[CD07]{cayton2008learning}
Lawrence Cayton and Sanjoy Dasgupta, \emph{A learning framework for nearest
  neighbor search}, Advances in Neural Information Processing Systems, 2007,
  pp.~233--240.

\bibitem[Chu96]{chung1996laplacians}
Fan~RK Chung, \emph{Laplacians of graphs and cheeger’s inequalities},
  Combinatorics, Paul Erdos is Eighty \textbf{2} (1996), no.~157-172, 13--2.

\bibitem[DKZ{\etalchar{+}}17]{khalil2017learning}
Hanjun Dai, Elias Khalil, Yuyu Zhang, Bistra Dilkina, and Le~Song,
  \emph{Learning combinatorial optimization algorithms over graphs}, Advances
  in Neural Information Processing Systems, 2017, pp.~6351--6361.

\bibitem[DS13]{dasgupta2013randomized}
Sanjoy Dasgupta and Kaushik Sinha, \emph{Randomized partition trees for exact
  nearest neighbor search}, Conference on Learning Theory, 2013, pp.~317--337.

\bibitem[DSN17]{dasgupta2017neural}
Sanjoy Dasgupta, Charles~F Stevens, and Saket Navlakha, \emph{A neural
  algorithm for a fundamental computing problem}, Science \textbf{358} (2017),
  no.~6364, 793--796.

\bibitem[GB10]{glorot2010}
Xavier Glorot and Yoshua Bengio, \emph{Understanding the difficulty of training
  deep feedforward neural networks}, International Conference on Artificial
  Intelligence and Statistics, 2010, pp.~249--256.

\bibitem[GLGP13]{gong2013iterative}
Yunchao Gong, Svetlana Lazebnik, Albert Gordo, and Florent Perronnin,
  \emph{Iterative quantization: A procrustean approach to learning binary codes
  for large-scale image retrieval}, IEEE Transactions on Pattern Analysis and
  Machine Intelligence \textbf{35} (2013), no.~12, 2916--2929.

\bibitem[IS15]{ioffe2015batch}
Sergey Ioffe and Christian Szegedy, \emph{Batch normalization: Accelerating
  deep network training by reducing internal covariate shift}, arXiv preprint
  arXiv:1502.03167 (2015).

\bibitem[JDJ17]{johnson2017billion}
Jeff Johnson, Matthijs Douze, and Herv{\'e} J{\'e}gou, \emph{Billion-scale
  similarity search with gpus}, arXiv preprint arXiv:1702.08734 (2017).

\bibitem[JDS11]{jegou2011product}
Herve J\'egou, Matthijs Douze, and Cordelia Schmid, \emph{Product quantization
  for nearest neighbor search}, IEEE transactions on pattern analysis and
  machine intelligence \textbf{33} (2011), no.~1, 117--128.

\bibitem[KB15]{adam2015}
Diederik Kingma and Jimmy Ba, \emph{Adam: A method for stochastic
  optimization}, International Conference for Learning Representations, 2015.

\bibitem[KBC{\etalchar{+}}18]{kraska2017case}
Tim Kraska, Alex Beutel, Ed~H Chi, Jeffrey Dean, and Neoklis Polyzotis,
  \emph{The case for learned index structures}, Proceedings of the 2018
  International Conference on Management of Data, ACM, 2018, pp.~489--504.

\bibitem[KS18]{keivani2018improved}
Omid Keivani and Kaushik Sinha, \emph{Improved nearest neighbor search using
  auxiliary information and priority functions}, International Conference on
  Machine Learning, 2018, pp.~2578--2586.

\bibitem[KSKW15]{kusner2015word}
Matt Kusner, Yu~Sun, Nicholas Kolkin, and Kilian Weinberger, \emph{From word
  embeddings to document distances}, International Conference on Machine
  Learning, 2015, pp.~957--966.

\bibitem[KZN08]{kumar2008good}
Neeraj Kumar, Li~Zhang, and Shree Nayar, \emph{What is a good nearest neighbors
  algorithm for finding similar patches in images?}, European conference on
  computer vision, Springer, 2008, pp.~364--378.

\bibitem[LCY{\etalchar{+}}17]{li2017losha}
Jinfeng Li, James Cheng, Fan Yang, Yuzhen Huang, Yunjian Zhao, Xiao Yan, and
  Ruihao Zhao, \emph{Losha: A general framework for scalable locality sensitive
  hashing}, Proceedings of the 40th International ACM SIGIR Conference on
  Research and Development in Information Retrieval, ACM, 2017, pp.~635--644.

\bibitem[LJW{\etalchar{+}}07]{lv2007multi}
Qin Lv, William Josephson, Zhe Wang, Moses Charikar, and Kai Li,
  \emph{Multi-probe lsh: efficient indexing for high-dimensional similarity
  search}, Proceedings of the 33rd international conference on Very large data
  bases, VLDB Endowment, 2007, pp.~950--961.

\bibitem[LLW{\etalchar{+}}15]{erin2015deep}
Venice~Erin Liong, Jiwen Lu, Gang Wang, Pierre Moulin, and Jie Zhou, \emph{Deep
  hashing for compact binary codes learning}, Proceedings of the IEEE
  conference on computer vision and pattern recognition, 2015, pp.~2475--2483.

\bibitem[LNC{\etalchar{+}}11]{li2011learning}
Zhen Li, Huazhong Ning, Liangliang Cao, Tong Zhang, Yihong Gong, and Thomas~S
  Huang, \emph{Learning to search efficiently in high dimensions}, Advances in
  Neural Information Processing Systems, 2011, pp.~1710--1718.

\bibitem[LV18]{lykouris2018competitive}
Thodoris Lykouris and Sergei Vassilvitskii, \emph{Competitive caching with
  machine learned advice}, International Conference on Machine Learning, 2018.

\bibitem[Mit18]{mitz2018model}
Michael Mitzenmacher, \emph{A model for learned bloom filters and optimizing by
  sandwiching}, Advances in Neural Information Processing Systems, 2018.

\bibitem[MMB17]{metzler2017learned}
Chris Metzler, Ali Mousavi, and Richard Baraniuk, \emph{Learned d-amp:
  Principled neural network based compressive image recovery}, Advances in
  Neural Information Processing Systems, 2017, pp.~1772--1783.

\bibitem[MPB15]{mousavi2015deep}
Ali Mousavi, Ankit~B Patel, and Richard~G Baraniuk, \emph{A deep learning
  approach to structured signal recovery}, Communication, Control, and
  Computing (Allerton), 2015 53rd Annual Allerton Conference on, IEEE, 2015,
  pp.~1336--1343.

\bibitem[MY18]{malkov2018efficient}
Yury~A Malkov and Dmitry~A Yashunin, \emph{Efficient and robust approximate
  nearest neighbor search using hierarchical navigable small world graphs},
  IEEE transactions on pattern analysis and machine intelligence (2018).

\bibitem[NCB17]{nidetecting}
Y~Ni, K~Chu, and J~Bradley, \emph{Detecting abuse at scale: Locality sensitive
  hashing at uber engineering}, 2017.

\bibitem[PSK18]{purohit2018improving}
Manish Purohit, Zoya Svitkina, and Ravi Kumar, \emph{Improving online
  algorithms via ml predictions}, Advances in Neural Information Processing
  Systems, 2018, pp.~9661--9670.

\bibitem[PSM14]{pennington2014glove}
Jeffrey Pennington, Richard Socher, and Christopher Manning, \emph{Glove:
  Global vectors for word representation}, Proceedings of the 2014 conference
  on empirical methods in natural language processing (EMNLP), 2014,
  pp.~1532--1543.

\bibitem[RG13]{ram2013space}
Parikshit Ram and Alexander Gray, \emph{Which space partitioning tree to use
  for search?}, Advances in Neural Information Processing Systems, 2013,
  pp.~656--664.

\bibitem[SDSJ19]{sablayrolles2018spreading}
Alexandre Sablayrolles, Matthijs Douze, Cordelia Schmid, and Herve J\'egou,
  \emph{Spreading vectors for similarity search}, International Conference on
  Learning Representations, 2019.

\bibitem[Spr91]{sproull1991refinements}
Robert~F Sproull, \emph{Refinements to nearest-neighbor searching
  ink-dimensional trees}, Algorithmica \textbf{6} (1991), no.~1-6, 579--589.

\bibitem[SS13]{sandersschulz2013}
Peter Sanders and Christian Schulz, \emph{{Think Locally, Act Globally: Highly
  Balanced Graph Partitioning}}, Proceedings of the 12th International
  Symposium on Experimental Algorithms (SEA'13), LNCS, vol. 7933, Springer,
  2013, pp.~164--175.

\bibitem[SWQ{\etalchar{+}}14]{sun2014srs}
Yifang Sun, Wei Wang, Jianbin Qin, Ying Zhang, and Xuemin Lin, \emph{Srs:
  solving c-approximate nearest neighbor queries in high dimensional euclidean
  space with a tiny index}, Proceedings of the VLDB Endowment \textbf{8}
  (2014), no.~1, 1--12.

\bibitem[WGS{\etalchar{+}}17]{wu2017multiscale}
Xiang Wu, Ruiqi Guo, Ananda~Theertha Suresh, Sanjiv Kumar, Daniel~N
  Holtmann-Rice, David Simcha, and Felix Yu, \emph{Multiscale quantization for
  fast similarity search}, Advances in Neural Information Processing Systems,
  2017, pp.~5745--5755.

\bibitem[WLKC16]{wang2016learning}
Jun Wang, Wei Liu, Sanjiv Kumar, and Shih-Fu Chang, \emph{Learning to hash for
  indexing big data - a survey}, Proceedings of the IEEE \textbf{104} (2016),
  no.~1, 34--57.

\bibitem[WSSJ14]{wang2014hashing}
Jingdong Wang, Heng~Tao Shen, Jingkuan Song, and Jianqiu Ji, \emph{Hashing for
  similarity search: A survey}, arXiv preprint arXiv:1408.2927 (2014).

\bibitem[ZZ17]{zhang2017embedjoin}
Haoyu Zhang and Qin Zhang, \emph{Embedjoin: Efficient edit similarity joins via
  embeddings}, Proceedings of the 23rd ACM SIGKDD International Conference on
  Knowledge Discovery and Data Mining, ACM, 2017, pp.~585--594.

\end{thebibliography}
\bibliographystyle{amsalpha}

\appendix

\section{Results for MNIST}
\label{appendix_mnist}

\begin{figure*}[h]
    \centering
    \subfloat[One level, $16$ bins]{{\includegraphics[width=0.50\textwidth]{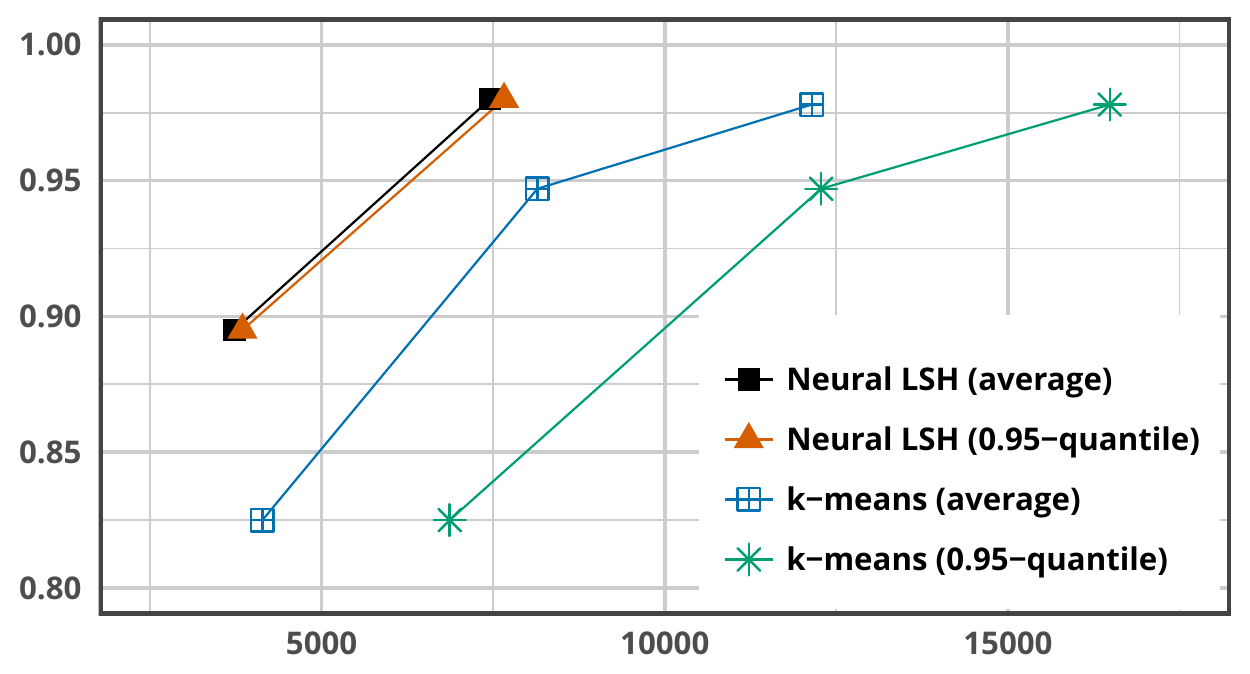}}}%
    \subfloat[Two levels, $16$ bins]{{\includegraphics[width=0.50\textwidth]{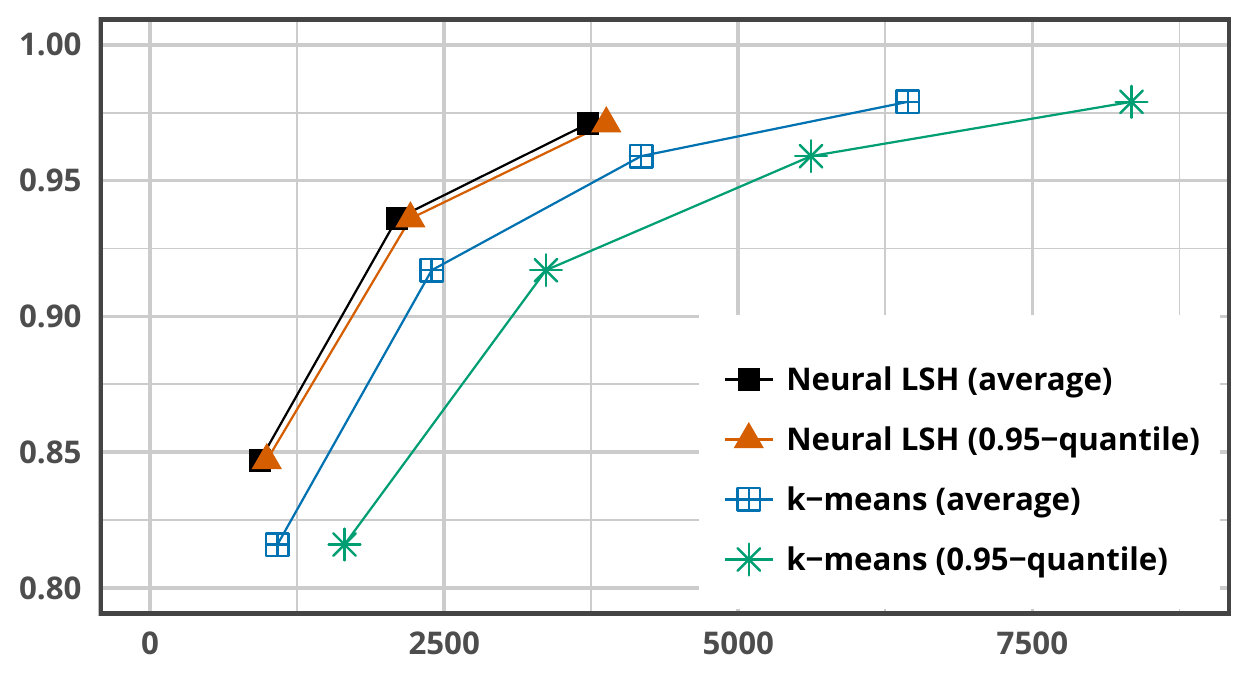}}}%

    \caption{MNIST, comparison of Neural LSH with $k$-means; x-axis -- number of candidates, y-axis -- $10$-NN accuracy}%
    \label{fig:mnist_kmeans}
\end{figure*}

\begin{figure*}[h]
    \centering
    \includegraphics[width=0.6\textwidth]{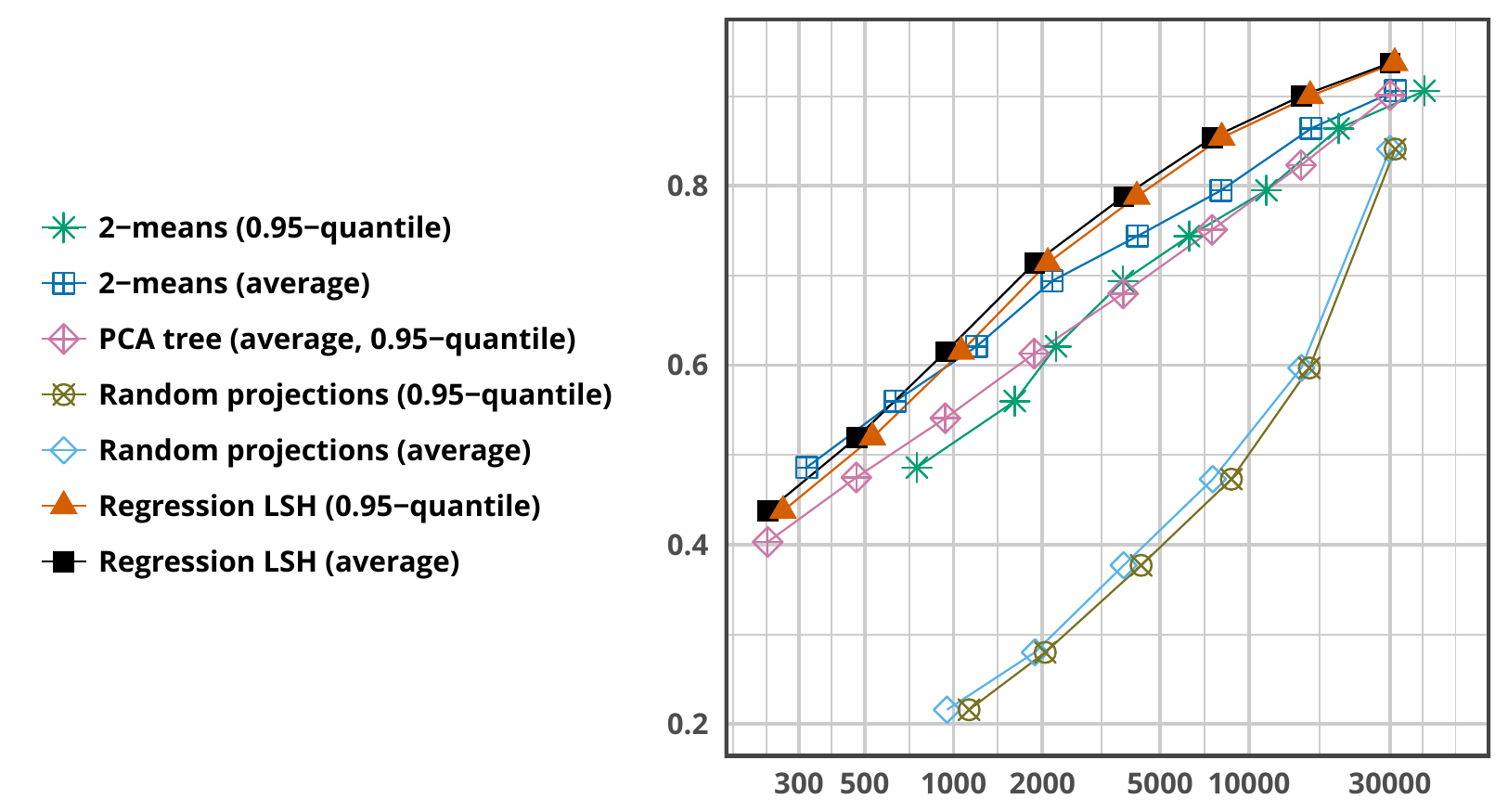}

    \caption{MNIST, comparison of trees built from hyperplanes; x-axis -- number of candidates, y-axis -- $10$-NN accuracy}%
    \label{fig:mnist_trees}
\end{figure*}

We include experimental results for the MNIST dataset, where all the experiments are performed exactly in the same way as for SIFT and GloVe. Consistent with the trend we observed for SIFT and GloVe, Neural LSH consistently outperforms $k$-means (see Figure~\ref{fig:mnist_kmeans}) both in terms of average number of candidates and especially in terms of the $0.95$-th quantiles. We also compare Regression LSH with recursive $2$-means,
as well as PCA tree and random projections (see Figure~\ref{fig:mnist_trees}), where Regression LSH consistently outperforms the other methods.

\section{Effect of Neural Catalyzer on Space Partitions}
\begin{figure*}[h]
    \centering
    \subfloat[GloVe, one level, $16$ bins]{{\includegraphics[width=0.50\textwidth]{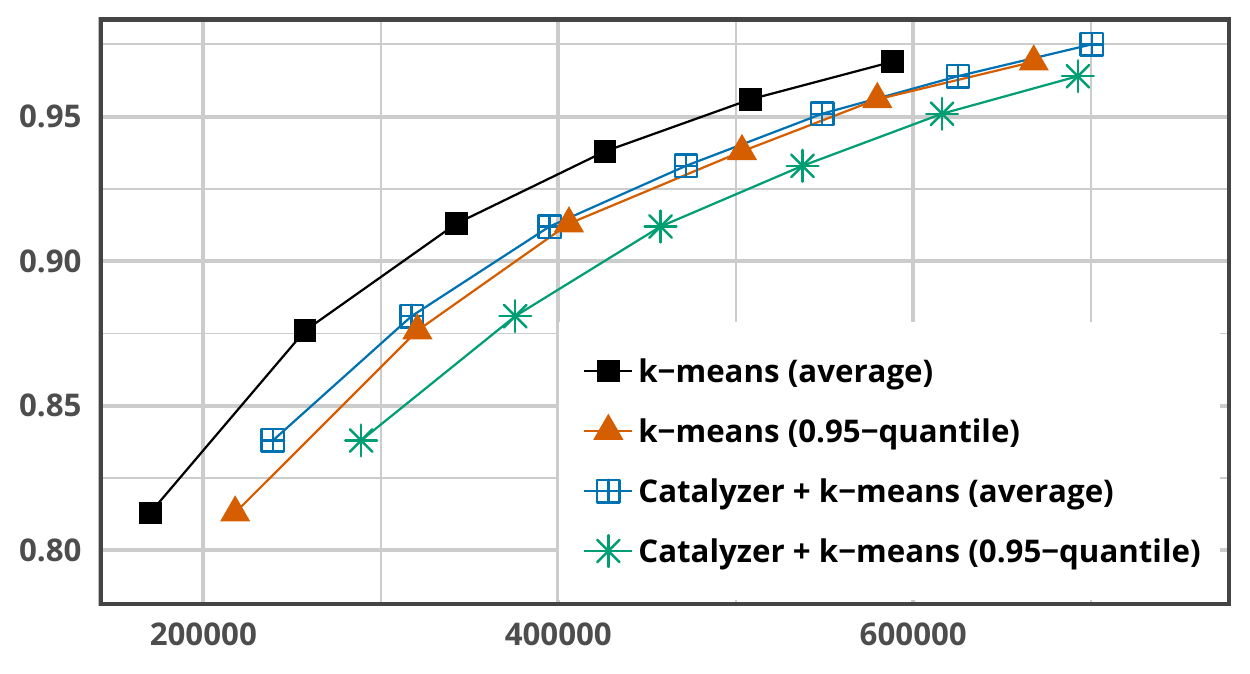}}}%
    \subfloat[SIFT, one level, $16$ bins]{{\includegraphics[width=0.50\textwidth]{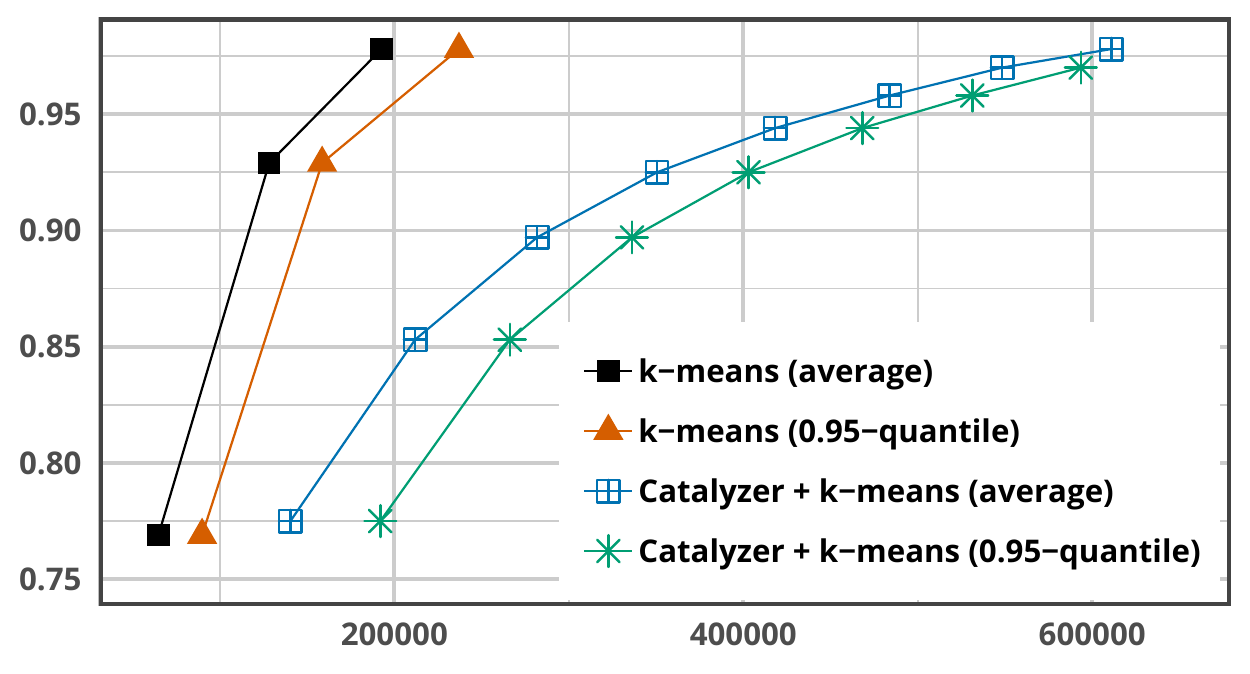}}}%
    \caption{Comparison of $k$-means and Catalyzer + $k$-means}
    \label{fig:catalyzer_plots}
\end{figure*}

In this section we compare vanilla $k$-means with $k$-means run after
applying a Neural Catalyzer map~\cite{sablayrolles2018spreading}. 
The goal is to check whether the Neural Catalyzer -- which is designed to boost up the performance of sketching methods for NNS by adjusting the input geometry -- could also improve the quality of space partitions for NNS.
See Figure~\ref{fig:catalyzer_plots} for the comparison on GloVe and SIFT with $16$ bins.
On both datasets (especially SIFT), Neural Catalyzer in fact degrades
the quality of the partitions.
We observed a similar trend for other numbers of bins than the setting reported here.
These findings support our observation that while both indexing and sketching for NNS can benefit from learning-based enhancements, they are fundamentally different approaches and require different specialized techniques.

%%%% Lonely figure floating in space %%%%

\begin{figure*}%
    \centering
    \subfloat[GloVe, one level, $16$ bins]{{\includegraphics[width=0.50\textwidth]{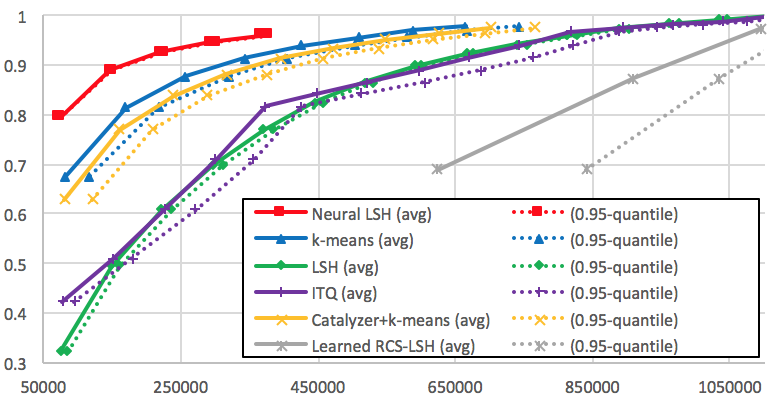}}}%
    \subfloat[SIFT, one level, $16$ bins]{{\includegraphics[width=0.50\textwidth]{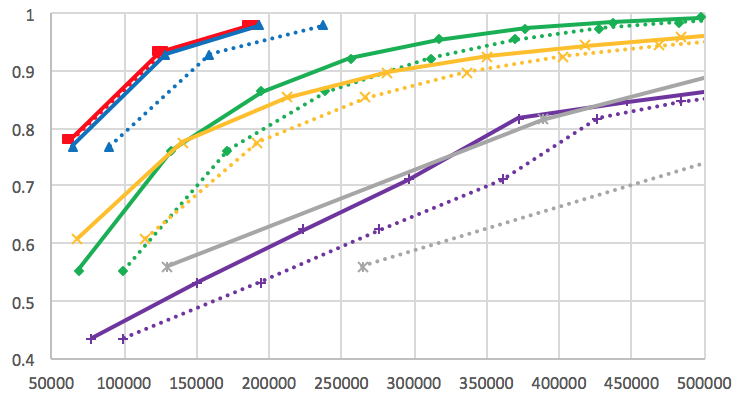}}}%

    \subfloat[GloVe, one level, $256$ bins]{{\includegraphics[width=0.50\textwidth]{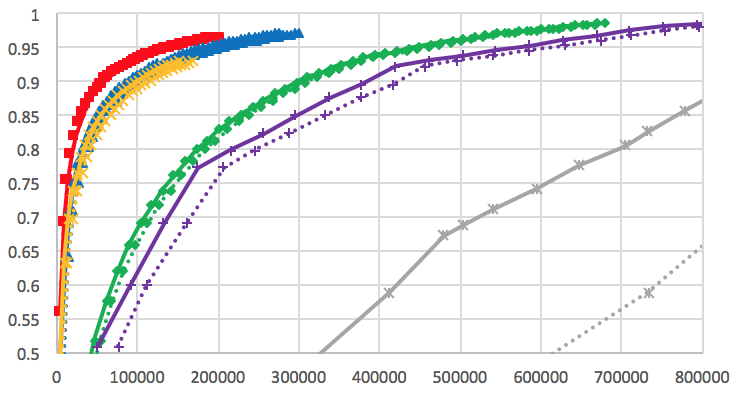}}}%
    \subfloat[SIFT, one level, $256$ bins]{{\includegraphics[width=0.50\textwidth]{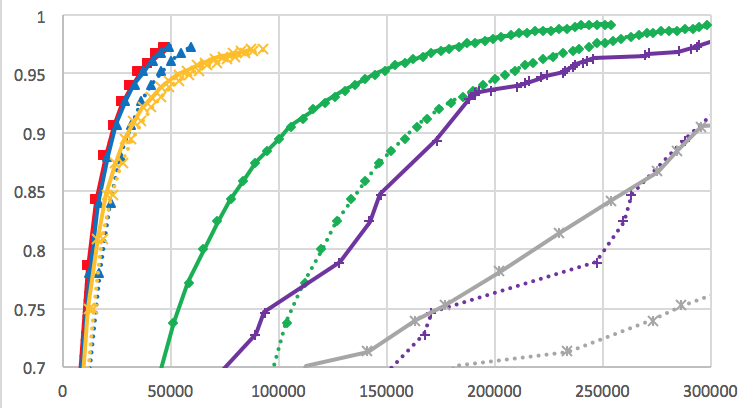}}}%
    
    \subfloat[GloVe, two levels, $16$ bins]{{\includegraphics[width=0.50\textwidth]{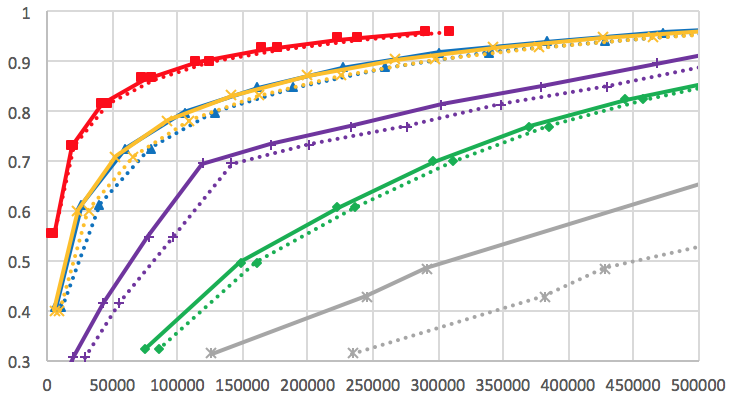}}}%
    \subfloat[SIFT, two levels, $16$ bins]{{\includegraphics[width=0.50\textwidth]{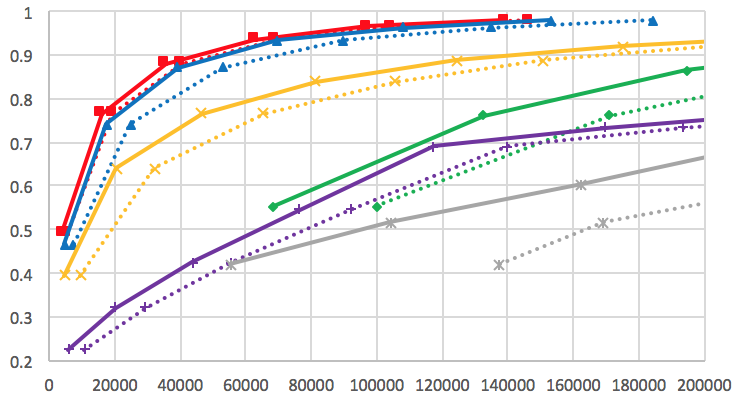}}}%

    \subfloat[GloVe, two levels, $256$ bins, $k$-means at 2nd level]{{\includegraphics[width=0.50\textwidth]{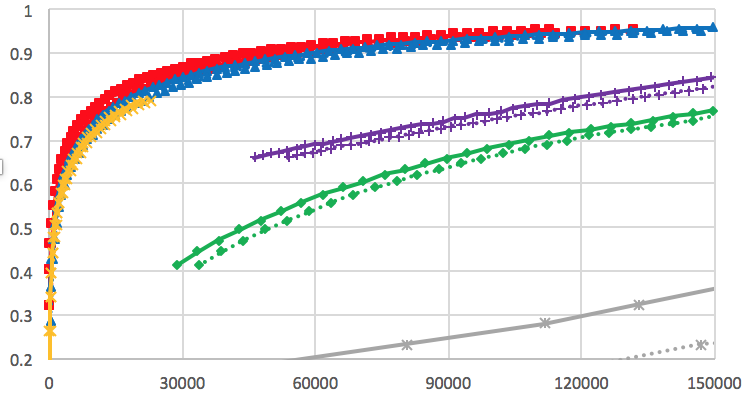}}}%
    \subfloat[SIFT, two levels, $256$ bins, $k$-means at 2nd level]{{\includegraphics[width=0.50\textwidth]{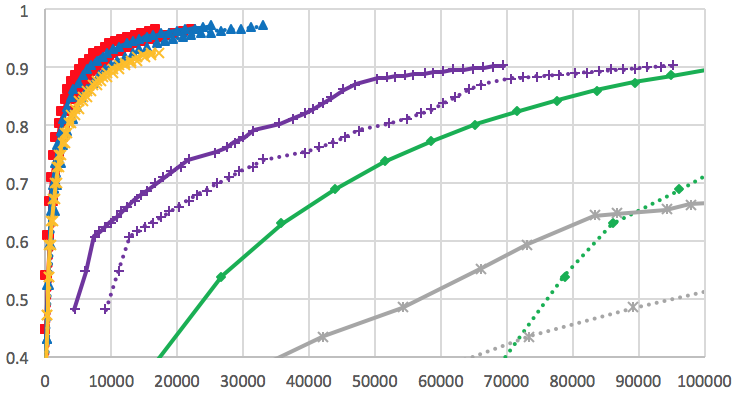}}}%
    \caption{Results from Figure~\ref{fig:kmeans_plots} with broader candidate and accuracy regimes. The ``Learned RCS-LSH'' baseline is the learned rectilinear cell structure locality sensitive hashing method of~\cite{cayton2008learning}.}%
    \label{fig:all_plots}
\end{figure*}

\section{Proof of Theorem~\ref{spectral_thm}}
\label{spectral_proof}

\begin{proof}
Consider an undirected graph $G = (V, E)$, where the set of vertices $V$ is $P$, and the (multi-)set of edges contains an edge $(p, p')$ for every $p' \in N_k(p)$.
The graph contains $n$ vertices and $kn$ edges, and some of the edges might be double (if $p' \in N_k(p)$ and $p \in N_k(p')$ at the same time).
Let $A_G$ be the symmetric adjacency matrix of $G$ normalized by $2kn$ (so that the sum of all the entries equals to $1$, thus giving a probability distribution over $P \times P$, which can be seen to be equal to $\mathcal{D}_{\mathrm{close}}$). The rows and columns of $A_G$ can naturally be indexed by the points of $P$.
Denote $\rho_G(p) = \sum_{p'} (A_G)_{p,p'}$. It is immediate to check that $\rho_G$ yields a distribution over $P$, which can be seen to be equal to $\mathcal{D}$.
Denote $D_G = \mathrm{diag}(\rho_G)$. Denote $L_G = D_G - A_G$ the Laplacian of $A_G$. Due to the equivalence of $\rho_G$ and $\mathcal{D}$ and $A_G$ and $\mathcal{D}_{\mathrm{close}}$, we have:
\begin{equation}
\frac{\alpha}{\beta} = \frac{\sum_{p, p' \in P} (A_G)_{p,p'} \cdot \|p - p'\|_2^2}{\sum_{p,p' \in P} \rho_G(p) \rho_G(p') \cdot \|p - p'\|_2^2}.
\end{equation}
By considering all possible coordinate projections and using additivity of $\|\cdot\|_2^2$ over coordinates, we conclude that there exists a coordinate $i^* \in [d]$ such that:
\begin{equation}
\label{eqeqeq1}
\frac{\sum_{p, p' \in P} (A_G)_{p,p'} \cdot (p_{i^*} - p'_{i^*})^2}{\sum_{p, p' \in P} \rho_G(p) \rho_G(p') \cdot (p_{i^*} - p_{i^*}')^2} \leq \frac{\alpha}{\beta}.
\end{equation}
Define a vector $y \in \Rbb^P$ by $y_p = p_{i^*}$. 
We now apply the following standard fact from spectral graph theory: If $A$ is the weighted adjacency matrix of a graph, and $L$ is its Laplacian matrix, then $x^tLx = \sum_{i,j=1}^nA_{ij}(x_i-x_j)^2$ for all $x\in\R^n$.
Thus the numerator of~(\ref{eqeqeq1}) becomes $y^t L_G y$.
For the denominator, consider the graph $H$ on $P$ in which every pair $p,p'$ is connected by an edge of weight $\rho_G(p)\rho_G(p')$.
\begin{CompactItemize}
\item Its weighted adjacency matrix $A_H$ is given by $(A_H)_{p,p'}=\rho_G(p)\rho_G(p')$ for $p\neq p'$, and with zeros on the diagonal. Thus $A_H=\rho_G\rho_G^t - D_G^2$ (recall that $D_G = \mathrm{diag}(\rho_G)$).
\item The degree of each node $p$ in $H$ equals $\rho_G(p)\sum_{p'\in P\setminus\{p\}}\rho_G(p') = \rho_G(p)-(\rho_G(p))^2$ (recall that $\sum_{p'\in P}\rho_G(p')=1$). Therefore the diagonal degree matrix of $H$ is $D_H=D_G-D_G^2$.
\end{CompactItemize}
Together, the Lapacian of $H$ is $L_H=D_H-A_H=D_G-\rho_G\rho_G^t$. Therefore the denominator of~(\ref{eqeqeq1}) becomes $y^t (D_G - \rho_G \rho_G^t) y$. Overall, we have:
$$
\frac{y^t L_G y}{y^t (D_G - \rho_G \rho_G^t) y}  \leq \frac{\alpha}{\beta} .
$$

Next, we define $\widetilde{y} = y - c\cdot\mathbf1$, where $\mathbf1$ is the all-$1$'s vector, and $c$ is the scalar $c=(y^t\rho_G)/(\mathbf1^t\rho_G)$.
This scalar is chosen to render $\widetilde{y}\perp\rho_G$.
Furthermore, since $\mathbf1$ is in the kernel of every Laplacian matrix, we have $L_G\widetilde{y}=L_Gy$ and $L_H\widetilde{y}=L_Hy$. Together, we get

$$
\frac{\widetilde{y}^t L_G \widetilde{y}}{\widetilde{y}^t D_G \widetilde{y}} = \frac{y^t L_G y}{y^t (D_G - \rho_G \rho_G^t) y} \leq \frac{\alpha}{\beta},
$$

Now by the Cheeger's inequality~\cite{chung1996laplacians}, we conclude that there exists a threshold $y_0 \in \Rbb$ such that:
\begin{equation}
\label{eqeqeq2}
\frac{\sum_{p_1, p_2: \widetilde{y}_{p_1} \leq y_0, \widetilde{y}_{p_2} > y_0} (A_G)_{p_1, p_2}}{\min\{\sum_{p : \widetilde{y}_p \leq y_0} \rho_G(p), \sum_{p : \widetilde{y}_p > y_0} \rho_G(p)\}} \leq \sqrt{2 \cdot \frac{\widetilde{y}^t L_G \widetilde{y}}{\widetilde{y}^t D_G \widetilde{y}}} \leq \sqrt{\frac{2 \alpha}{\beta}}.
\end{equation}
One can trace back all the definitions and observe that the set $\{p \in P \colon \widetilde{y}_p \leq y_0\}$ is induced by an (axis-aligned) hyperplane,
and the left-hand side of~(\ref{eqeqeq2}) is nothing else but the left-hand side of~(\ref{eqeqeq3}).
\end{proof}

%\subsubsection*{Acknowledgments}
%Use unnumbered third level headings for the acknowledgments. All
%acknowledgments, including those to funding agencies, go at the end of the paper.

\end{document}